\documentclass[sigconf,screen,authorversion,nonacm]{acmart} 
\AtBeginDocument{%
  }
    
\setcopyright{acmcopyright}
\copyrightyear{2024}
\acmYear{2024}
\acmDOI{XXXXXXX.XXXXXXX}

\acmJournal{TELO}
\acmVolume{37}
\acmNumber{4}
\acmArticle{111}
\acmMonth{8}

\usepackage{amsmath,amsfonts}
\usepackage{algorithm}
\usepackage{algorithmicx}
\usepackage{algpseudocode}
\usepackage{array}
\usepackage{textcomp}
\usepackage{stfloats}
\usepackage{url}
\usepackage{verbatim}
\usepackage{graphicx}
\usepackage{fancyvrb}
\usepackage{fvextra}
\usepackage{comment}
\usepackage{booktabs} 
\usepackage{xcolor}
\usepackage{placeins}
\usepackage{soul}

\usepackage{graphicx}
\usepackage{booktabs,balance}
\usepackage{dsfont}
\usepackage{comment}
\usepackage{subfigure}
\usepackage{svg}
\usepackage[frozencache,cachedir=.]{minted}
\usepackage{multirow}

\usepackage{tikz}
\usetikzlibrary{fit,calc}
\newcommand*{\tikzmk}[1]{\tikz[remember picture,overlay,] \node (#1) {};\ignorespaces}
\newcommand{\boxit}[1]{\tikz[remember picture,overlay]{\node[yshift=3pt,fill=#1,opacity=.25,fit={(A)($(B)+(-.1\linewidth,.8\baselineskip)$)}] {};}\ignorespaces}
\colorlet{myyellow}{yellow!40}
\colorlet{mypink}{red!40}
\colorlet{mycyan}{cyan!60}

\begin{document}



\title[In-the-loop HPO for Automated Design of Heuristics]{In-the-loop Hyper-Parameter Optimization for LLM-Based Automated Design of Heuristics}




\author{Niki van Stein}
\email{n.van.stein@liacs.leidenuniv.nl}
\orcid{0000-0002-0013-7969}
\affiliation{%
  \institution{LIACS, Leiden University}
  \streetaddress{Niels Bohrweg 1}
  \city{Leiden}
  \country{Netherlands}
  \postcode{NL-2333}
}

\author{Diederick Vermetten}
\email{d.vermetten@liacs.leidenuniv.nl}
\orcid{0000-0003-3040-7162}
\affiliation{%
  \institution{LIACS, Leiden University}
  \streetaddress{Niels Bohrweg 1}
  \city{Leiden}
  \country{Netherlands}
  \postcode{NL-2333}
}

\author{Thomas B{\"a}ck}
\email{t.h.w.baeck@liacs.leidenuniv.nl}
\orcid{0000-0001-6768-1478}
\affiliation{%
  \institution{LIACS, Leiden University}
  \streetaddress{Niels Bohrweg 1}
  \city{Leiden}
  \country{Netherlands}
  \postcode{NL-2333}
}

\renewcommand{\shortauthors}{van Stein et al.}

\begin{abstract}
Large Language Models (LLMs) have shown great potential in automatically generating and optimizing (meta)heuristics, making them valuable tools in heuristic optimization tasks. However, LLMs are generally inefficient when it comes to fine-tuning hyper-parameters of the generated algorithms, often requiring excessive queries that lead to high computational and financial costs. This paper presents a novel hybrid approach, LLaMEA-HPO, which integrates the open source LLaMEA (Large Language Model Evolutionary Algorithm) framework with a Hyper-Parameter Optimization (HPO) procedure in the loop. By offloading hyper-parameter tuning to an HPO procedure, the LLaMEA-HPO framework allows the LLM to focus on generating novel algorithmic structures, reducing the number of required LLM queries and improving the overall efficiency of the optimization process. 

We empirically validate the proposed hybrid framework on benchmark problems, including Online Bin Packing, Black-Box Optimization, and the Traveling Salesperson Problem. Our results demonstrate that LLaMEA-HPO achieves superior or comparable performance compared to existing LLM-driven frameworks while significantly reducing computational costs. This work highlights the importance of separating algorithmic innovation and structural code search from parameter tuning in LLM-driven code optimization and offers a scalable approach to improve the efficiency and effectiveness of LLM-based code generation.
\end{abstract}



\received{01 October 2024}

\maketitle


\section{Introduction}

Large Language Models (LLMs) have demonstrated remarkable capabilities in generating, optimizing, and refining algorithms autonomously \cite{FunSearch2024,fei2024eoh,vanstein2024llamealargelanguagemodel}, making them powerful tools for various heuristic optimization tasks. The intersection of LLMs with evolutionary algorithms has led to promising advancements in the automatic design of optimization algorithms and heuristics, as frameworks such as FunSearch \cite{FunSearch2024}, Evolution of Heuristics (EoH) \cite{fei2024eoh} and Large Language Model Evolutionary Algorithm (LLaMEA) \cite{vanstein2024llamealargelanguagemodel} have shown. 
These models can generate entire algorithmic structures, providing innovative solutions to complex optimization problems. However, one significant limitation of LLM-driven algorithm generation and optimization is the relatively high financial and computational budget required.
Either specialized hardware is required to load large enough LLM models locally, or funding is required to use third-party APIs such as OpenAI's ChatGPT. 
In addition, 
it was observed already that parts of the evolutionary search using LLMs focuses purely on the fine-tuning of hyper-parameters of the generated (meta)heuristics
\cite{vanstein2024llamealargelanguagemodel}. The primary tasks which LLMs are trained for and excel in, are the generation of plausible natural language, and thereby also plausible Python code, but they are generally speaking not trained for generating good numerical hyper-parameter settings, and using an LLM to only perform hyper-parameter tuning is very costly.

This paper introduces a hybridization of the open source LLaMEA framework with a dedicated Hyper-Parameter Optimization (HPO) procedure, to handle hyper-parameter tuning. By offloading the task of tuning numerical parameters to an HPO tool, the LLaMEA-HPO framework allows the LLM to focus on more creative tasks, such as generating novel algorithmic structures and control flows. This approach significantly reduces the number of LLM queries needed and enhances the performance of the evolutionary process while lowering financial and computational costs.

The contributions of this work are threefold:
\begin{itemize}
    \item We propose \emph{LLaMEA-HPO}, a novel framework that integrates LLM-driven algorithm design with SMAC-based \cite{smac3} hyper-parameter optimization, resulting in more efficient code generation and optimization.
    \item We empirically demonstrate that delegating hyper-parameter tuning to SMAC significantly reduces the LLM query budget and computational costs while achieving state-of-the-art performance across various benchmark problems, including Online Bin Packing, Black-Box Optimization, and Traveling Salesperson Problem.
    \item We provide insights into the balance between algorithmic creativity and parameter tuning in LLM-driven frameworks, offering recommendations for future research and practical applications in computationally expensive domains.
\end{itemize}

The remainder of this paper is organized as follows: Section \ref{sec:related} discusses related work in LLM-driven evolutionary computing and HPO. Section \ref{sec:methodology} introduces the LLaMEA-HPO methodology, explaining the hybridization of the LLaMEA framework with SMAC for hyper-parameter tuning. Section \ref{sec:experiments} outlines the experimental setup, including the benchmarks used for evaluation. Section \ref{sec:results} presents the results and discusses these, and Section \ref{sec:conclusions} concludes with recommendations for future work.

\section{Related Work}
\label{sec:related}

The combination 
of Large Language Models (LLMs) and optimization has led to novel advancements in automated heuristic and algorithm generation. LLMs have demonstrated remarkable capabilities in generating, refining, and optimizing algorithms autonomously. 

LLMs have emerged as effective tools in automating the design and optimization of code. The \emph{FunSearch} framework leverages LLMs to explore function spaces, producing programs for combinatorial optimization tasks such as the cap-set problem and bin packing \cite{FunSearch2024}. This approach employs an island-based Evolutionary Algorithm (EA) to ensure diversity, while the LLM iteratively generates and refines solutions. Similarly, \emph{Algorithm Evolution using Large Language Models (AEL)} \cite{liu2023algorithm} uses LLMs to evolve optimization heuristics, applying crossover and mutation operators to code snippets. The \emph{Evolution of Heuristics (EoH)} approach \cite{fei2024eoh} extends this by treating each algorithm as a population candidate, with LLMs evolving these through heuristic-based mutations. Although these methods show promise, particularly in small-scale instances of problems like the Traveling Salesperson Problem (TSP), their scalability to more complex domains such as continuous optimization remains limited.

Recent research by Zhang et al. \cite{zhang2024understanding} explored the integration of LLMs with evolutionary program search (EPS) methods for automated heuristic design (AHD) in depth. Their work provides a comprehensive benchmark across several LLM-based EPS methods, aiming to assess the impact of combining LLMs with evolutionary search strategies. The study highlights the importance of evolutionary search in improving the quality of heuristics generated by LLMs, as standalone LLMs often fail to achieve competitive results. They also point out the high variance in performance across different LLMs and problem types, suggesting that further improvements in LLM-based AHD require more sophisticated search strategies and prompt designs.

Another recent contribution to this field is the LLaMEA (Large Language Model Evolutionary Algorithm) framework \cite{vanstein2024llamealargelanguagemodel}, which integrates LLMs within an evolutionary loop to automatically generate and optimize metaheuristic algorithms for solving continuous, unconstrained single-objective optimization problems. 
LLaMEA iteratively evolves algorithms by leveraging LLMs to generate, mutate, and select candidates based on performance metrics and code feedback, such as runtime error tracebacks. By integrating LLMs with benchmarking tools like IOHexperimenter~\cite{IOHexperimenter}, LLaMEA can systematically evaluate generated algorithms and optimize them for black-box optimization tasks. This framework has shown that LLMs can generate novel metaheuristics that outperform state-of-the-art methods, such as Covariance Matrix Adaptation Evolution Strategy (CMA-ES) and Differential Evolution (DE), on black-box optimization benchmarks.
LLaMEA addresses the limitations of previous evolution frameworks with LLMs by automating the generation and evaluation of more complex algorithms. Unlike approaches such as EoH, which are typically limited to generating small heuristic functions, LLaMEA can generate complete optimization algorithms that consist of multiple interacting components, including classes with variables and functions. The framework also supports iterative refinement based on detailed performance feedback, enabling it to adapt and optimize for a wide range of continuous optimization tasks.

Despite the significant advancements that frameworks like LLaMEA and EoH have introduced in LLM-driven algorithm optimization, they are not without limitations. One of the primary challenges in deploying LLMs within evolutionary computing is the high computational and financial cost associated with LLM inference. Most state-of-the-art approaches rely on third-party API services for querying LLMs, which incurs substantial costs, particularly when the LLM is repeatedly queried in an evolutionary loop. Alternatively, deploying and running LLM models in-house requires significant computational resources and infrastructure, especially for larger models, further adding to the practical burden. This computational overhead makes it difficult to scale these approaches to larger or more complex optimization tasks without significant investment in computational power or financial resources.
Another notable limitation is that LLMs, when employed for algorithm generation, tend to spend a significant fraction of LLM-calls on the tuning of the hyper-parameters of the generated algorithms rather than proposing fundamentally novel control flows or algorithmic components \cite{vanstein2024llamealargelanguagemodel}. 
While hyper-parameter optimization (see e.g.~\cite{baratchi2024automated} for a comprehensive overview of the state-of-the-art) is an essential aspect of fine-tuning algorithm performance, the reliance on LLMs to merely adjust existing parameters rather than innovate in terms of new algorithmic structures reduces the potential for groundbreaking improvements. This behavior is likely influenced by the training data of LLMs, which is abundant with examples of traditional optimization techniques and their refinements but lacks substantial exposure to novel algorithmic innovation. As a result, many of the generated solutions are either slight modifications of well-known algorithms or tuned versions of previously generated candidates, limiting the diversity and novelty of solutions.


\section{Methodology}\label{sec:methodology}

These gaps in the current state of LLM-driven evolutionary computing methods serve as the main motivation for the development of LLaMEA-HPO. By integrating Hyper-Parameter Optimization (HPO) techniques within the LLM-driven framework, LLaMEA-HPO aims to offload the task of hyper-parameter tuning from the LLM, allowing it to focus on higher-level tasks such as generating novel control flows and algorithmic components. In this hybrid framework, the LLM generates candidate algorithms, while an HPO tool is employed to optimize their hyper-parameters in the loop (i.e., for each generated metaheuristic), improving efficiency and reducing the computational costs associated with repeated LLM queries for minor tuning tasks. This separation of concerns not only alleviates some of the financial and computational burden but also enhances the LLM’s ability to explore new algorithmic structures, leading to more diverse and innovative solutions.

In this section, we first provide a detailed explanation of the \emph{LLaMEA} framework, designed to integrate LLMs within an evolutionary computation loop for metaheuristic generation. Next we explain how the hyper-parameter optimization inside the proposed LLaMEA-HPO algorithms works and last but not least how these components interact together.

\subsection{The LLaMEA Framework}

The \textbf{LLaMEA} framework operates in the following stages \cite{vanstein2024llamealargelanguagemodel}:
\begin{enumerate}
    \item \textbf{Initial Generation:} The process begins with an initial candidate algorithm generated by the LLM given a detailed task prompt and one example algorithm. These algorithms are represented as Python code snippets with a short description.
    
    \item \textbf{Evaluation:} The candidate algorithm is evaluated based on its performance on a given set of tasks, such as the optimization over different black-box functions. This performance is quantified using a predefined fitness function, typically based on solution quality, computational efficiency, or convergence speed.
    
    \item \textbf{Selection:} The best-performing candidate so-far is selected for the next iteration. In the original LLaMEA framework this is a (1+1) evolutionary algorithm, meaning we have only one solution at a given time and we select the best-so-far as the parent individual to continue the search.
    
    \item \textbf{Mutation:} The selected candidate undergoes a mutation-like operation performed by the LLM based on a feedback prompt. This prompt notifies the LLM of the attained fitness value of the best-so-far algorithm and asks the LLM to refine the algorithm in order to improve it. 
    
    \item \textbf{Termination:} This process is repeated for a predefined number of generations or until a termination criterion is met, such as reaching a performance threshold.
\end{enumerate}


While traditional evolutionary algorithms rely on predefined crossover and mutation operators, LLaMEA leverages the LLM to generate novel variations of solutions by simply asking it to mutate (improve) the best-so-far candidate. 
By incorporating LLM-driven creativity into the evolutionary process to generate and optimize code, LLaMEA is able to explore a broad search space of algorithmic designs, leading to potentially innovative and effective algorithms for a wide range of problems.

LLaMEA distinguishes itself from Evolution of Heuristics (EoH) in several factors:
\begin{itemize}
    \item LLaMEA uses a self-debugging procedure, feeding back error traces from occuring run-time errors to improve the solutions generated by the LLM.
    \item LLaMEA generated complete Python classes, including class parameters and functions. This gives a lot of flexibility compared to EoH, which can only produce single functions with predetermined in and output variables.
    \item LLaMEA uses a (1+1) evolutionary algorithm, which is known to be a competitive strategy especially under small budget constraints. In addition, it is also a very simple evolutionary algorithm, making it possible to analyse the evolutionary runs straightforwardly.
\end{itemize}

\subsection{Hyper-parameter Optimization}

It is well known that an algorithm's parameterization can have a significant impact on its behaviour and corresponding performance, both in the context of optimization and in machine learning~\cite{feurer2019hyperparameter, bischl2023hyperparameter, hoos2012automated}. As such, hyper-parameter optimization (or algorithm configuration) is often a critical step when developing or benchmarking a new algorithm. 
A wide variety of automated algorithm configuration tools have been developed over the last decades
\cite{baratchi2024automated}.

One could argue that algorithm configuration is already encompassed in the broader question of automated algorithm design. Indeed, determining the setting of an algorithm's parameters is implicitly included in the design space through which e.g.~LLaMEA can search, but this would be a rather inefficient use of prompts. 
As such, 
As an alternative, we augment the approach of LLaMEA by including a separate hyper-parameter optimization stage before evaluating the LLM's proposed algorithm. 

For this hyper-parameter optimization step, we make use of SMAC3~\cite{smac3}. Since our goal generally is to find an algorithm which works well on a set of problem instances, rather than a single one, we take a portion of our available instances as the training set that SMAC can use during its search. After a given budget of algorithm runs is exhausted, SMAC's final incumbent solution is evaluated in the full set of problem instances to achieve a consistent score for each algorithm proposed by the LLM.

In order to run this hyper-parameter optimization procedure effectively, we need to define a valid search space for the algorithm's hyper-parameters. In our case, we generate this by modifying the LLaMEA prompt to provide an example configuration space (in the format of the ConfigSpace package~\cite{configspace}) and formatting the requested response to include this configuration space for the proposed algorithm.


\subsection{LLaMEA-HPO}

\begin{figure*}[!ht]
    \centering
    \includegraphics[width=\textwidth, trim=5mm 5mm 5mm 5mm,clip]{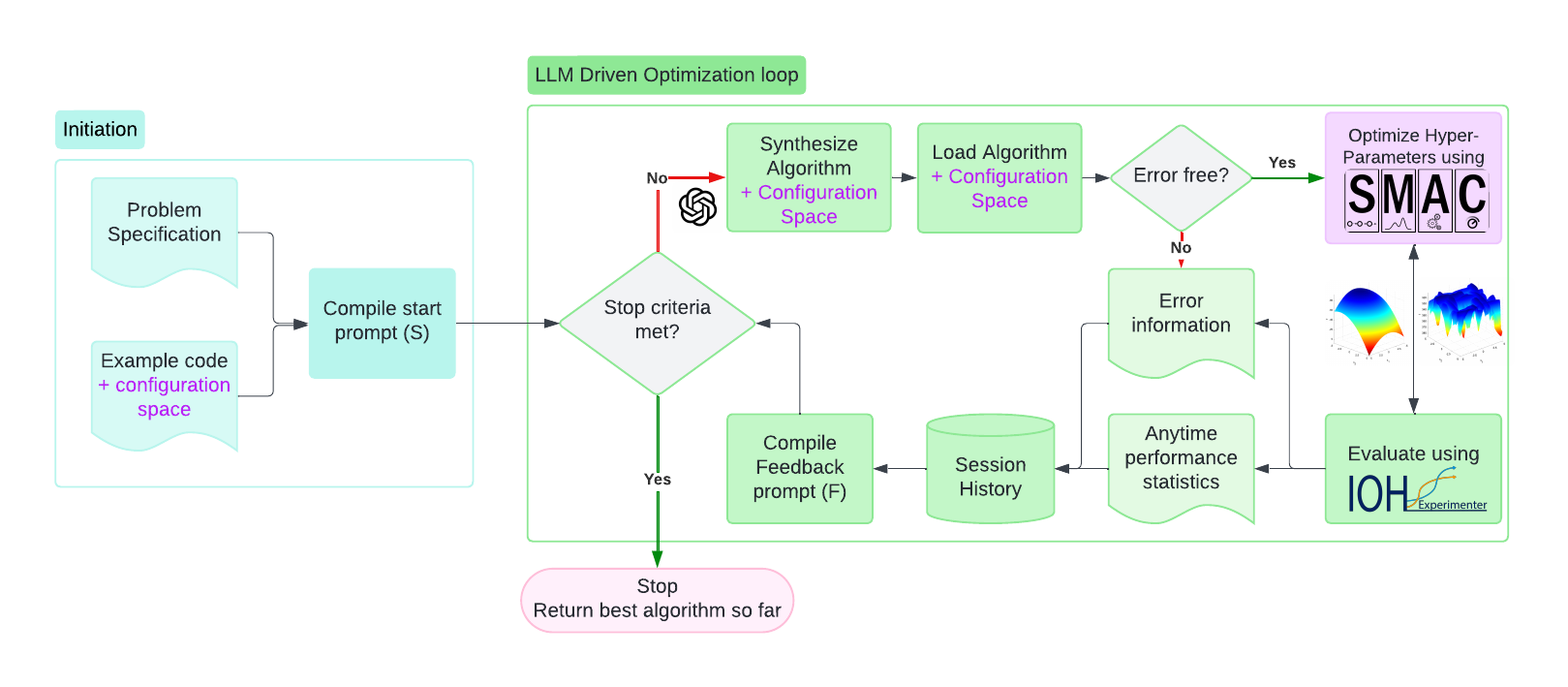}
    \caption{The summary of the proposed LLM driven algorithm design with Hyper-parameter optimization framework LLaMEA-HPO. Marked in {\color{violet} violet} are the components related to HPO. 
    \label{fig:framework}}
\end{figure*}

The proposed LLaMEA-HPO procedure builds upon the existing open-source LLaMEA framework\footnote{\url{https://zenodo.org/records/13268663}} by integrating a specialized Hyper-Parameter Optimization (HPO) component, in this case SMAC, to improve the efficiency of LLM-driven evolutionary code generation. As depicted in Figure \ref{fig:framework} and highlighted in Algorithm \ref{alg:LLaMEAHPO}, the primary innovation in this hybrid approach is the offloading of hyper-parameter tuning from the LLM to an HPO method. This hybridization allows the LLM to focus on generating and mutating algorithmic structures while the HPO component handles parameter tuning, reducing computational overhead and the number of costly LLM queries.

The LLaMEA-HPO procedure follows the key steps outlined below:

\textbf{1. Initial Task Prompt and Algorithm Generation:}  
The process begins by providing the LLM with a detailed task prompt $S$, which specifies the optimization task to be solved and an example of a candidate algorithm (one-shot prompting). In this step, the LLM generates an initial algorithm and, crucially, also provides a configuration space for its hyper-parameters in the form of a Python dictionary. This dictionary is essential for the subsequent HPO step.
The task prompt $S$ follows the following structure:

\begin{tiny}
\begin{Verbatim}[breaklines=true, frame=single, label=Detailed task prompt $S$]
<specific task context>
An example of such code is as follows:
```python
<initial example code>
```
The parameters of the algorithm you provide will be optimized by SMAC. Provide the configuration space as Python dictionary of all the hyper-parameters of the provided algorithm.
An example configuration space is as follows:
```python
<example configuration space dictionary>
```
Provide a novel algorithm and its configuration space to solve
this task. 
Give the response in the format:
# Name: <name of the algorithm>
# Code: <code>
# Configspace: <configuration space>
\end{Verbatim}
\end{tiny}

\textbf{2. Hyper-Parameter Optimization (HPO):}  
Once the initial algorithm and its configuration space are generated, the HPO procedure is invoked. Here, we use SMAC to optimize the hyper-parameters of the generated algorithm. SMAC searches the configuration space (generated by the LLM), using a subset of problem instances to evaluate candidate configurations. After a set number of iterations, the optimized algorithm is returned. This optimization step significantly reduces the burden on the LLM, which is typically less efficient at fine-tuning numerical parameters and more efficient in language (code) tasks.

\textbf{3. Algorithm Evaluation:}  
The LLaMEA-HPO framework evaluates the optimized algorithm using a fitness function. The evaluation typically involves solving a set of problem instances (e.g., Online Bin Packing, BBOB, TSP) and measuring the algorithm’s performance based on average solution quality, computational efficiency, or other relevant metrics. Errors encountered during execution are captured and fed back to the LLM in a self-debugging procedure. Error-free solutions are considered for further refinement if they improve over the best-so-far.

\textbf{4. Feedback Loop and Mutation:}  
After each evaluation, a feedback prompt $F$ is constructed. This prompt contains the name, code, configuration space, and performance metrics of the best-performing algorithm with optimized hyper-parameters so far. The LLM is then queried again, using this feedback to mutate the algorithm. The mutation process involves refining the control flows, algorithmic structures, or logic of the code while leaving the hyper-parameters unchanged. The resulting offspring algorithm and its (updated) configuration space are passed to the HPO module for further tuning. The feedback prompt $F$ follows the following structure:

\begin{tiny}
\begin{Verbatim}[breaklines=true,frame=single, label=Feedback prompt template $F$]
<Task prompt S>
<List of previously generated algorithm names and their score>
<selected algorithm to refine (full code) score and optimal hyper-parameters>
Either refine or redesign to improve the algorithm and provide both the code and a new configuration space.
\end{Verbatim}
\end{tiny}
\textbf{5. Iteration and Termination:}  
The LLaMEA-HPO process repeats this loop of LLM-driven mutation and HPO optimization until a specified stopping criterion is met (e.g., a maximum number of LLM queries or achieving a predefined performance threshold). At the end of the procedure, the best-performing algorithm and its optimized hyper-parameters are returned.

\textbf{Advantages of Hybridization:}  
By incorporating HPO into the LLaMEA framework, the proposed method achieves two key benefits:
1) \textit{Reduction of the number of LLM Queries:} The LLM is reserved for generating novel algorithmic structures, avoiding costly LLM prompts for minor hyper-parameter tuning tasks. This not only reduces the financial costs of using LLMs but also improves the efficiency of the evolutionary process.
2) \textit{Enhanced Solution Quality:} The integration of HPO ensures that the hyper-parameters of the algorithms are well-tuned for the given optimization tasks, possibily leading to better performance on benchmarks.

The steps highlighted in purple in Figure \ref{fig:framework} represent the additional components introduced by this hybrid approach, specifically the generation of the configuration space, the application of HPO, and the optimized feedback loop that includes refined algorithms and hyper-parameters. Algorithm \ref{alg:LLaMEAHPO} formalizes these steps, showing how LLM and HPO components interact in each iteration to produce increasingly optimized solutions.
For illustration, we have highlighted the modifications from the original LLaMEA algorithm (see \cite{vanstein2024llamealargelanguagemodel}) in light blue.

\begin{algorithm*}
\caption{LLaMEA-HPO\label{alg:LLaMEAHPO}}
\begin{algorithmic}[1]
\State $T \gets ; T_s \gets ; f_i \gets ; f \gets$
\Comment{Given LLM prompt budget $T$, HPO budget $T_s$, instance-based evaluation function $f_i$, final evaluation function (problem) $f$}
\State $S \gets $ task-prompt
\Comment{Task description prompt}
\State $F_0 \gets $ task-feedback-prompt
\Comment{Feedback prompt after each iteration}
\State $t \gets 0$
\State $a_t, c_t \gets LLM(S)$
\Comment{Initialize by generating first parent program and configuration space}
\tikzmk{A} 

\State $a_t^* \gets HPO(f_i, a_t, c_t, T_s)$ \Comment{Perform HPO on $a_t$ using $c_t$, budget $T_s$ and evaluation function $f_i$}

\tikzmk{B}
\boxit{mycyan}
\State $(y_t, \sigma_t, e_t) \gets f(a_t^*)$
\Comment{Evaluate mean quality and std.-dev.~of first program on $f$ and catch errors if occurring}
\State $a_b \gets a_t^*; y_b \gets y_t; \sigma_b \gets \sigma_t; e_b \gets e_t$
\Comment{Remember best-so-far}
\While{$t < T$}     
    \Comment{Budget not exhausted}
    \Comment{Construct new prompt, using best-so-far algorithm}
    \State $F \gets (S, ((\text{name}(a_0), y_0),\ldots,(\text{name}(a_t), y_t)), (a_b, y_b, \sigma_b, e_b), F_0)$ 
    
    \State $a_{t+1}, c_{t+1} \gets LLM(F)$ 
    \Comment{Generate offspring algorithm and configuration space by mutation}
\tikzmk{A} 

    \State $a_{t+1}^* \gets HPO(f_i, a_{t+1}, c_{t+1}, T_s)$
    \Comment{Perform HPO on $a_{t+1}$}

\tikzmk{B} 
\boxit{mycyan}
    \State $(y_{t+1}, \sigma_{t+1}, e_{t+1}) \gets f(a_{t+1}^*)$
    \Comment{Evaluate optimized offspring algorithm, catch errors}
    \If{$e_{t+1} \neq \emptyset$} 
        $y_{t+1} = 0$ \Comment{Errors occurred}
    \EndIf
    \If{$y_{t+1} \geq y_t$} 
        $a_b \gets a_{t+1}^*$; $y_b \gets y_{t+1}$; $\sigma_b \gets \sigma_{t+1}$; $e_b \gets e_{t+1}$ 
        \Comment{Update best}
    \EndIf
    \State $t \gets t+1$ \Comment{Increase LLM prompt counter}
\EndWhile
\State \textbf{return} $a_b, y_b$ 
\Comment{Return best (configured) algorithm and its quality}
\end{algorithmic}
\end{algorithm*}

\section{Experimental Setup}
\label{sec:experiments}

In this work, we utilize three heuristic optimization problems as benchmarks: \emph{Online Bin Packing}, \emph{Traveling Salesperson Problem} (TSP) using Guided Local Search (GLS) and \emph{black-box optimization} using a large variety of noiseless continuous functions. These problems represent challenging combinatorial and black-box optimization scenarios often used to evaluate the performance of (meta)heuristic-based approaches.

In each of these experiments we look at the fitness over two different measures of (computational) cost.
The first is the number of LLM prompts or queries required by the LLM-driven optimization framework, and the second measure is the number of full benchmark evaluations. With a full benchmark evaluation we mean the validation of one generated (meta)heuristic on all training instances of the specific problem we are trying to solve. The proposed LLaMEA-HPO procedure uses instance-based hyper-parameter optimization inside the LLM-driven optimization loop,  meaning that for each LLM query we have several of these benchmark evaluations, depending on the HPO budget we use. To explain this further by a concrete example, for the black-box optimization benchmark we are using $24 \cdot 3 \cdot 3 = 216$ problem instances during the optimization loop ($24$ functions, $3$ instances per function and $3$ random seeds), evaluating an generated metaheuristic on all these $216$ problems would count as $1$ complete benchmark evaluation. Since we are using $2000$ \textbf{instance} evaluations as the budget of our hyper-parameter optimization procedure, we use per iteration $\frac{2000}{216} = 9.25$ full benchmark evaluations. Depending on the cost of doing the evaluation and querying the LLM for solving real-world problems, one might prefer more LLM queries versus less evaluation time or the other way around. In this work we primarily aim to reduce the number of LLM queries required, but for fairness and completeness we also report the evaluation time required.

\subsection{Online Bin Packing} 
The Online Bin Packing problem involves a sequence of items with varying sizes that must be packed into the minimum number of fixed-sized bins with capacity $C$. The challenge is to assign each item to a bin without knowing the future sequence of items, thus requiring an efficient, real-time decision-making process. We use here the online scenario from \cite{seiden2002online}.  

\subsubsection*{Evaluation}
The evaluation instances used during the LLaMEA evolution loop are $5$ Weibull instances of size 5k with a capacity of $100$ (as in \cite{FunSearch2024}). The fitness value is set as the average $\frac{lb}{n}$ over all instances, where $lb$ is the lower bound of the optimal number of bins \cite{martello1990lower} and $n$ the number of bins required by the heuristic solution. 

\subsubsection*{Baselines}
We evaluate our proposed approach here against the Evolution of Heuristics (EoH) approach, where the authors also use this benchmark with the exact same settings. In addition we also test our proposed approach against a vanilla LLaMEA algorithm using $1000$ evaluations and LLM queries as budget. For specifics about their hyper-parameter settings for the EoH algorithm we refer to their work \cite{fei2024eoh}. In turn, EoH already shows to outperform FunSearch \cite{FunSearch2024} and human hand-crafted heuristics. Both frameworks use the gpt-4o-2024-05-13 LLM \cite{openai2023chatgpt4o}.

\subsection{Black-Box Continuous Optimization using the BBOB Suite}
The second benchmark used in this study is the Black-Box Continuous Optimization, specifically utilizing the 24 noiseless benchmark functions from the well-established BBOB suite~\cite{hansen2009real}. These functions cover a wide range of problem characteristics, enabling a thorough evaluation of an algorithm's performance across diverse optimization landscapes.

The BBOB suite includes separable functions, functions with moderate and high conditioning, as well as unimodal and multimodal functions with varying global structure. This diversity challenges algorithms to effectively balance exploration and exploitation while navigating different topologies and constraints.

We use the BBOB benchmark function suite \cite{hansen2009real}
within IOHexperimenter~\cite{IOHexperimenter} in the same experimental setup as performed in \cite{vanstein2024llamealargelanguagemodel}.
In our experiments we also set the dimensionality of the optimization problems to $d=5$.

\subsubsection*{Evaluation}

To comprehensively evaluate the performance of the generated algorithms across the full set of BBOB benchmark functions, we employ an \emph{anytime performance measure}. This measure assesses the performance of an optimization algorithm over the entire budget, rather than focusing solely on the final objective function value. Specifically, we use the normalized \emph{Area Over the Convergence Curve} ($AOCC$) \cite{lopez2014automatically} also used in the experimental setup of  \cite{vanstein2024llamealargelanguagemodel}. 
The formulation for the $AOCC$ is provided in Equation (\ref{eq:AOCC}).
\begin{equation} \label{eq:AOCC}
\hspace{-1.15em}
    \textit{AOCC}(\mathbf{y}_{a,f}) = \frac{1}{B} \sum_{i=1}^{B} \left( 1-\frac{\min(\max(y_i, \textit{lb}), \textit{ub}) - \textit{lb}}{\textit{ub} - \textit{lb}} \right)
\end{equation}

In this equation, $\mathbf{y}_{a,f}$ represents the series of the best log-scaled precision values, i.e., the differences $\log(y_i - f^*)$ between the observed function value $y_i$ and the global minimum $f^*$ for the corresponding function, obtained throughout the optimization process of algorithm $a$ on the test function $f$. Here, $y_i$ is the $i$-th element of the sequence, $B=10\,000$ denotes the budget, and $\textit{lb}$ and $\textit{ub}$ are the lower and upper bounds defining the range of function values of interest. The standard bounds applied are $\textit{lb}=10^{-8}$ and $\textit{ub}=10^2$.

In line with established practices~\cite{hansen2022anytime}, these precision values undergo logarithmic scaling prior to the $AOCC$ computation. The $AOCC$ can also be interpreted as the area under the \emph{Empirical Cumulative Distribution Function} (ECDF) curve, considering an infinite number of target values within the specified bounds~\cite{eafecdf}.

To aggregate the $AOCC$ scores over all 24 benchmark functions in the BBOB suite, we compute the mean across functions and their instances. For an algorithm $a$, the aggregation is given by
\begin{equation}
    \textit{AOCC}(a) = \frac{1}{3 \cdot 24} \sum_{i=1}^{24} \sum_{j=1}^3 \textit{AOCC}(\mathbf{y}_{a, f_{ij}}) \; ,
\end{equation}
where $f_{ij}$ refers to the $j$-th instance of the $i$-th benchmark function.

Finally, the overall mean $AOCC$ score, averaged over $k = 3$  independent runs of algorithm $a$ across all BBOB functions, is used as the feedback to the LLM in the next stage of optimization. This aggregated score is treated as the best-so-far solution if an improvement is found. In formal terms, the performance metric $f(a)$ used in Algorithm \ref{alg:LLaMEAHPO} is defined as
\begin{equation}\label{eqn:f}
    f(a) = \frac{1}{k} \sum_{i=1}^k \textit{AOCC}(a) \; .
\end{equation}

Additionally, any runtime or compilation errors encountered during the validation phase are considered in the feedback to the language model. In the case of critical errors that prevent execution, the mean $AOCC$ score is set to the minimum value, which is zero.

\subsubsection*{Baselines}
Here we compare against the open-source LLaMEA \cite{vanstein2024llamealargelanguagemodel} framework and the EoH \cite{fei2024eoh} algorithm for generating metaheuristics. EoH is adapted slightly, as explained in \cite{vanstein2024llamealargelanguagemodel} in order for it to work in this task. EoH uses the LLM gpt-4o-2024-05-13 \cite{openai2023chatgpt4o} and the LLaMEA baseline uses gpt-4-turbo-2024-04-09 \cite{openai2023chatgpt4turbo}, LLaMEA with GPT-4o results are also available in \cite{vanstein2024llamealargelanguagemodel} but this version of LLaMEA was performing less stable. Note that GPT-4-Turbo is more expensive to use than GPT-4o (at the time of writing), GPT-4o is only $20\%$ of the costs of GPT-4-turbo. Since we propose to offload numerical optimization from the LLM evolution loop in order to preserve costs and have an as-efficient as possible LLM-driven search algorithm, we decided to use the cheaper GPT-4o as the only model we consider for the proposed approach in this work.

\subsection{Traveling Salesperson Problem (TSP) with Guided Local Search} The TSP \cite{matai2010traveling} is a well-known NP-hard problem where the goal is to determine the shortest possible route that visits each city exactly once and returns to the origin city. For this benchmark, we incorporate Guided Local Search (GLS), which enhances the basic local search technique by penalizing frequently used edges in sub-optimal tours. GLS is designed to escape local optima by dynamically adjusting penalties based on the frequency of edge usage, encouraging the exploration of new paths. The to be generated heuristic is a function that determines this edge costs to guide the local search algorithm. 

\subsubsection*{Evaluation}
For evaluation we use 64 TSP100 instances (TSP instances with $100$ locations). The locations are randomly sampled from $[0,1]^2$ as in \cite{kool2018attention}. As fitness score we calculate the average gap from the optimal solutions (in percentages), the optimal solutions are generated by Concorde \cite{applegate2009certification}. 

\subsubsection*{Baselines}
We use a similar comparison as presented in \cite{fei2024eoh}, where a wide range of the state-of-the-art methods is compared. This includes; Graph Convolutional Network (GCN) method for TSP \cite{joshi2019efficient}.
 Attention Model (AM) \cite{kool2018attention}, GLS \cite{voudouris1999guided}, the vanilla version of guided local search for TSP, EBGLS \cite{shi2018eb}, KGLS \cite{arnold2019knowledge}, GNNGLS \cite{hudson2021graph} and NeuralGLS \cite{sui2024neuralgls}. 
 In addition we compare to the heuristics found by the EoH algorithm for three independent runs (available open-source on the EoH repository \cite{fei2024eoh}), where we took the results from these runs after $100$ LLM prompts, and after the full $2000$ prompts used in these runs. Final evaluation of the generated heuristics is on a set of $3000$ instances, $1000$ per problem size for sizes $20$, $50$ and $100$. In addition, we evaluate these heuristics on all Euclidean TSPlib \cite{reinelt1991tsplib} problems. Both LLaMEA-HPO and EoH frameworks use the gpt-4o-2024-05-13 \cite{openai2023chatgpt4o} LLM.
 
\section{Results and Discussion}
\label{sec:results}

\subsection{Online Bin Packing}

\begin{figure*}[!htbp]
 \centering
  \includegraphics[width=.49\linewidth,trim=0mm 0mm 0mm 0mm,clip]{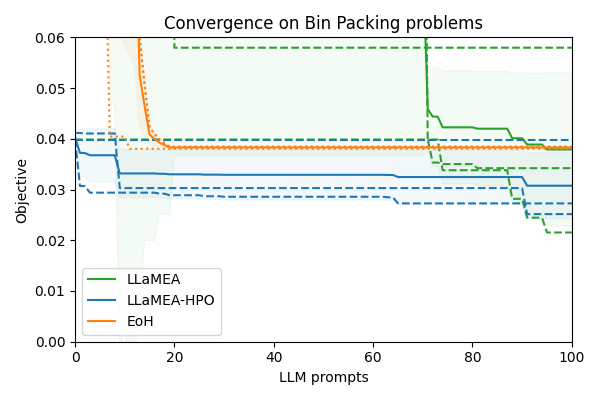}
   \includegraphics[width=.49\linewidth,trim=0mm 0mm 0mm 0mm,clip]{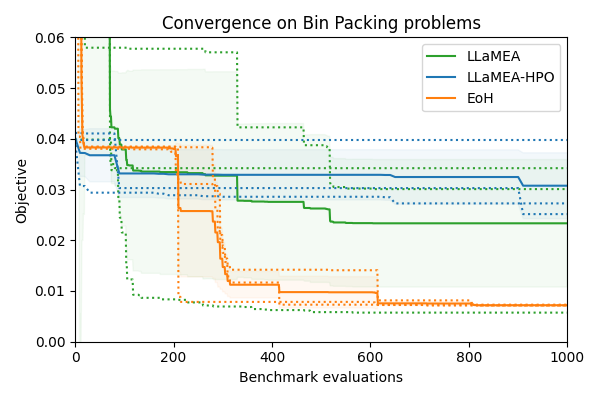}
 
 \caption{Convergence curves (lower is better) of the LLaMEA-HPO algorithm and the EoH algorithm on Online Bin Packing Problems (BP). Individual runs are denoted by dotted lines, the average convergence by a solid line and the standard deviation by the shaded area. The left plot shows the convergence with the number of LLM prompts used on the X-axis, and the right plot shows the convergence with the number of full benchmark evaluations used (including HPO evaluations). \label{fig:bp}
 }
 \Description[Several lines showing the convergence of EoH and LLaMEA-HPO on Bin Packing.]
 
\end{figure*}

Figure \ref{fig:bp} illustrates the convergence behavior of the LLaMEA-HPO algorithm compared to the EoH and vanilla LLaMEA algorithms on the Online Bin Packing Problem (BP). The individual runs are represented by dotted lines, while the solid lines depict the average convergence over these runs. Standard deviation is indicated by the shaded areas. On the left side of the figure, the convergence is shown with respect to the number of LLM prompts used, while the right side displays convergence in terms of the number of full benchmark evaluations. A full benchmark evaluation means that all the training instances (in the case of BP, there are $5$ Weibull 5k instances with a capacity of $100$) are evaluated exactly once. Since in LLaMEA-HPO we perform additional instance based hyper-parameter optimization, each LLM iteration in LLaMEA-HPO is using $10$ full benchmark evaluations. The HPO part of LLaMEA-HPO uses a total of $40$ instance evaluations, with the minimum number of instances for comparing is set to $1$ and the maximum set to $4$. The LLaMEA-HPO shows a more efficient convergence in terms of LLM queries used for the total budget of $100$ LLM queries. On the other hand, the proposed LLaMEA-HPO algorithm does not reach the same fitness value as EoH or LLaMEA after a full run of $1000$ benchmark evaluations (and in case of EoH that means $2000$ LLM prompts while for LLaMEA it means $1000$ LLM prompts, as for every solution EoH is quarrying the LLM twice). So while LLaMEA-HPO shows potential when we have a small LLM budget, it requires additional benchmark evaluations for the Online Bin Packing problem. It is also interesting to note that the vanilla LLaMEA version shows large differences between the runs, this can probably be explained by the (1+1) strategy it is using, larger populations would reduce this instability. LLaMEA-HPO does not seem to suffer from that, likely due to the hyper-parameter optimization part.

\subsection{BBOB 5D}

\begin{figure*}[!htbp]
 \centering
  \includegraphics[width=.49\linewidth,trim=0mm 0mm 0mm 0mm,clip]{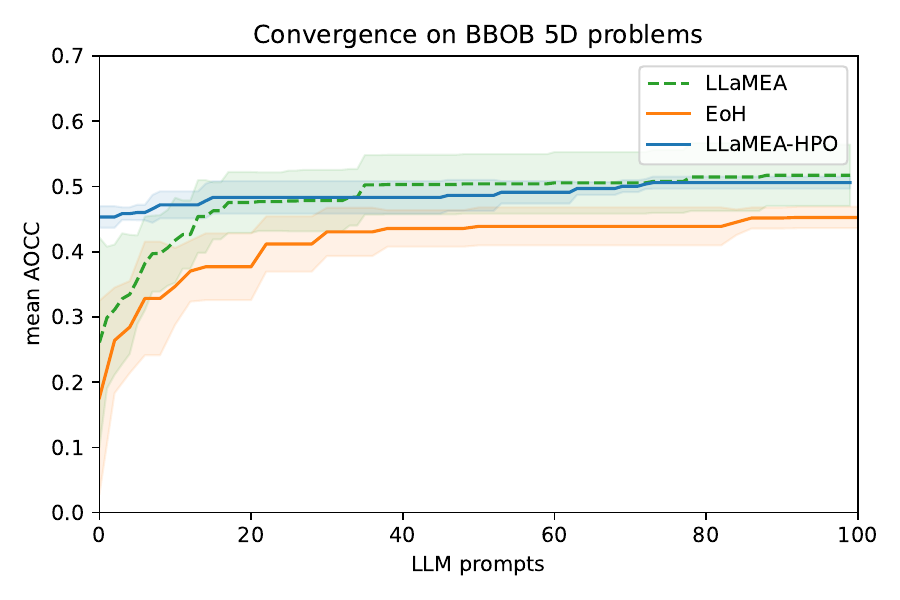}
   \includegraphics[width=.49\linewidth,trim=0mm 0mm 0mm 0mm,clip]{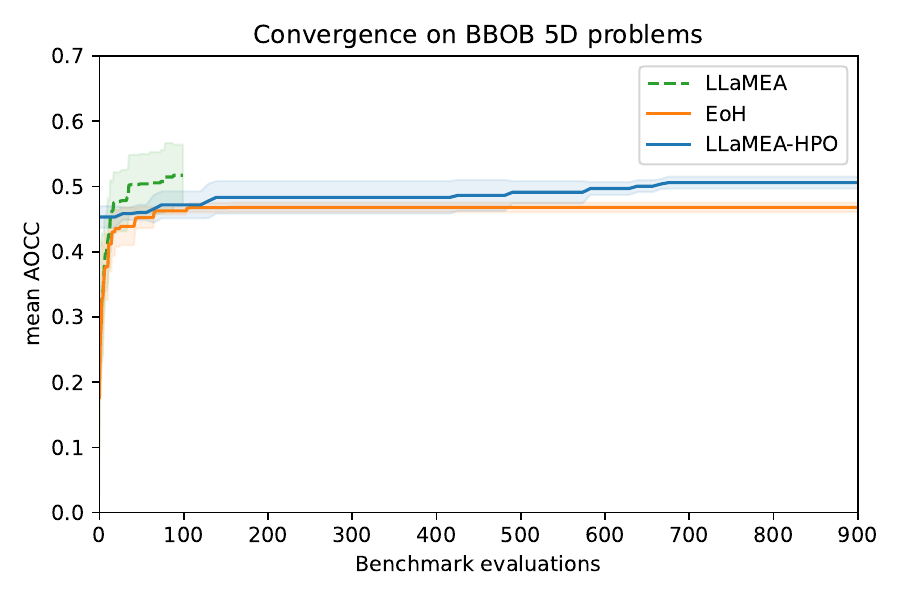}
 
 \caption{Convergence curves of the mean AOCC (higher is better) of the LLaMEA-HPO algorithm and the EoH algorithm on $5D$ BBOB problems. The average convergence over $5$ individual runs are denoted by a solid line and the standard deviation by the shaded area. The left plot shows the convergence with the number of LLM prompts used on the X-axis, and the right plot shows the convergence with the number of full benchmark evaluations used (including HPO instance evaluations). \label{fig:bbob}
 }
 \Description[Several lines showing the convergence of EoH and LLaMEA-HPO on Bin Packing.]
 
\end{figure*}

Figure \ref{fig:bbob} presents the convergence curves of the mean AOCC (Area Over the Convergence Curve) for the LLaMEA-HPO algorithm and the EoH algorithm on $5d$ BBOB problems. The results show that LLaMEA-HPO consistently achieves better performance than EoH, with higher mean AOCC values, as reflected by both the number of LLM prompts and full benchmark evaluations. In this use-case, we gave the HPO part a budget of $2000$ instance evaluations and as such the proposed LLaMEA-HPO algorithm requires $2000 / 216 = 9.25$ full benchmark evaluations (consisting of $9 \cdot 24 = 216$ instances) per LLM query. EoH used $1\,800$ LLM queries for the same $900$ benchmark evaluations (right side of Figure \ref{fig:bbob}). Note that the competing LLaMEA algorithm without HPO is using a $5$ times more expensive LLM (GPT-4-Turbo). It is interesting to note that the proposed method achieves state-of-the-art performance after a very few LLM queries, and it finds better algorithms after the first $20$ queries than the EoH algorithm is able to find after $1\,800$ LLM queries.

Next we compare the performance of the algorithms found by LLaMEA-HPO before and after their hyper-parameter optimization procedure to see the extend in which HPO matters.

\begin{figure}[!htbp]
 \centering
  \includegraphics[width=.6\linewidth,trim=0mm 0mm 0mm 0mm,clip]{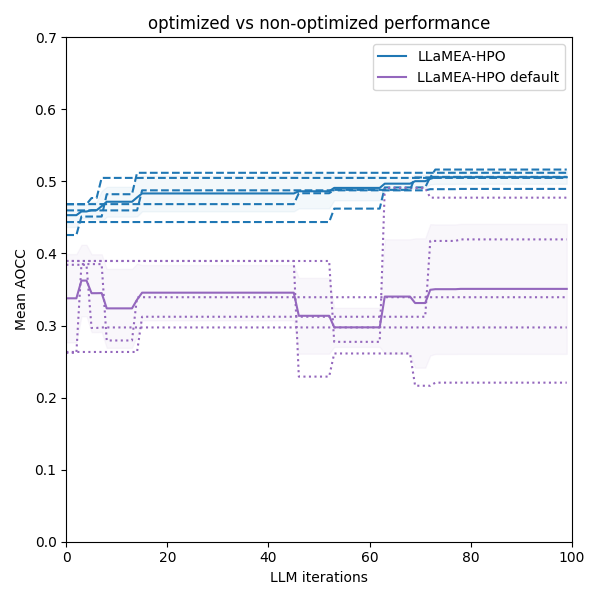}
 
 \caption{Convergence curves for LLaMEA-HPO (dotted lines are individual runs) and the non-hyper-parameter optimized algorithms as baseline (LLaMEA-HPO default).\label{fig:bbobdefault}
 }
 \Description[Several lines showing the convergence of different algorithms]
 
\end{figure}

In Figure \ref{fig:bbobdefault} we can observe that the algorithms generated by the LLM with generated default hyper-parameters have a mean AOCC between $0.25$ and $0.50$ (purple lines), their optimized counterparts, denoted in blue, show average AOCC scores between $0.45$ and $0.52$. On average the difference between non-optimized and optimized hyper-parameters is around $0.2$ in AOCC score, which is a significant difference. This underscores the effectiveness of the proposed hybridization. It also shows that much of the LLM-driven optimization in the baselines LLaMEA (without HPO) and EoH is actually the tuning of the different hyper-parameters, as also earlier observed in code diffs between parents and offspring in \cite{vanstein2024llamealargelanguagemodel}.

 

Next, we compare the resulting metaheuristic algorithms from three independent runs of LLaMEA-HPO with the best metaheuristic algorithm discovered in the original LLaMEA paper \cite{vanstein2024llamealargelanguagemodel}, called ``ERADS\_QuantumFluxUltraRefined''. For this we use the \emph{Glicko2} score procedure \cite{glickman2012example}. In this procedure there are a number of games per BBOB function, in this case $200$. In every round, for every function of the dataset, each pair of algorithms competes. This competition samples a random budget value for the provided runtime (with a maximum of $10\,000$). Whichever algorithm has the better function value at this budget wins the game. Then, from these games, the glico2-rating is used to determine the ranking.

\begin{figure}[!htbp]
 \centering
  \includegraphics[width=.98\linewidth,trim=0mm 0mm 0mm 0mm,clip]{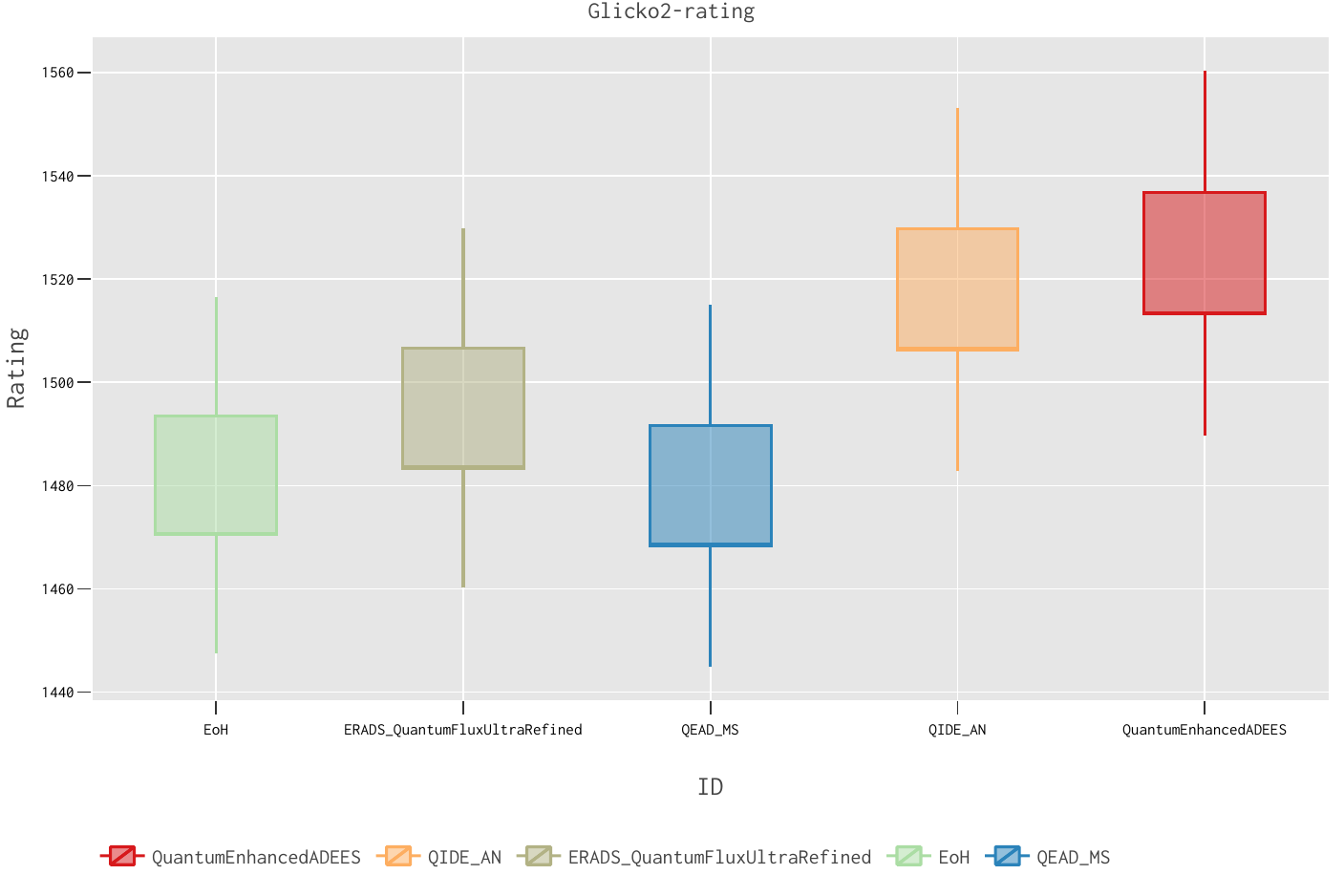}
 \caption{Glicko-2 \cite{glickman2012example} rating (higher is better) of each ``QEAD\_MS'', ``QIDE\_AN'' and ``QuantumEnhancedADEES'', the three hyper-parameter tuned algorithms resulting from three independent runs of LLaMEA-HPO, versus the original LLaMEA generated algorithm ``ERADS\_QuantumFluxUltraRefined'' and the best EoH generated algorithm (denoted ``EoH''). Glicko2 uses 200 matches, picking a random run at the specified budget of $10\,000$ per BBOB function between pairs of algorithms.\label{fig:glicko2}
 }
 \Description[Glicko2 rating per algorithm]
 
\end{figure}

\begin{table*}[!ht]
\centering
\caption{Results of the Glicko2 procedure using $200$ matches per BBOB function between each algorithm.}
\label{tab:glicko2}
\resizebox{\textwidth}{!}{
\begin{tabular}{rlllllllll}
  \hline
 & ID & Rating & Deviation & Volatility & Games & Win & Draw & Loss & Lag \\ 
  \hline
1 & QuantumEnhancedADEES & 1525 & 11.8 & 0.0237 & 19200 & 6126 & 8675 & 4399 & 0 \\ 
  2 & QIDE\_AN & 1518 & 11.7 & 0.0230 & 19200 & 5514 & 8669 & 5017 & 0 \\ 
  3 & ERADS\_QuantumFluxUltraRefined & 1495 & 11.6 & 0.0230 & 19200 & 5200 & 8825 & 5175 & 0 \\ 
  4 & EoH & 1482 & 11.5 & 0.0223 & 19200 & 5031 & 8010 & 6159 & 0 \\ 
  5 & QEAD\_MS & 1480 & 11.7 & 0.0233 & 19200 & 5074 & 7931 & 6195 & 0 \\ 
   \hline
\end{tabular}
}
\end{table*}

The resulting Glicko-2 ratings for the three  algorithms generated by LLaMEA-HPO, the best algorithm found by EoH and the previously found "ERADS\_QUantumFluxUltraRefined" algorithm are shown in Figure \ref{fig:glicko2} and Table \ref{tab:glicko2}. It is clear that the algorithms are performing comparable on $5D$ BBOB, with the new ``QuantumEnhancedADEES'' as slightly better performing algorithm with regard to the fixed-budget setting and final function value found. Overall, also given their AOCC score, they perform similarly in any-time performance. Source codes of these algorithms, all code and results are available on our Zenodo repository \cite{anonymous_2024_13834123}.


 

\subsection{Traveling Sales Person}

\begin{figure*}[!htbp]
 \centering
  \includegraphics[width=.49\linewidth,trim=0mm 0mm 0mm 0mm,clip]{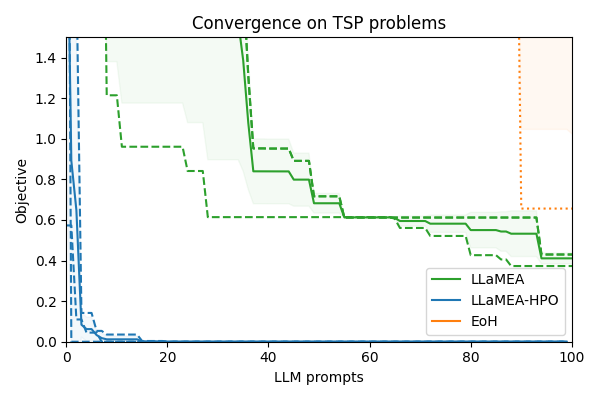}
   \includegraphics[width=.49\linewidth,trim=0mm 0mm 0mm 0mm,clip]{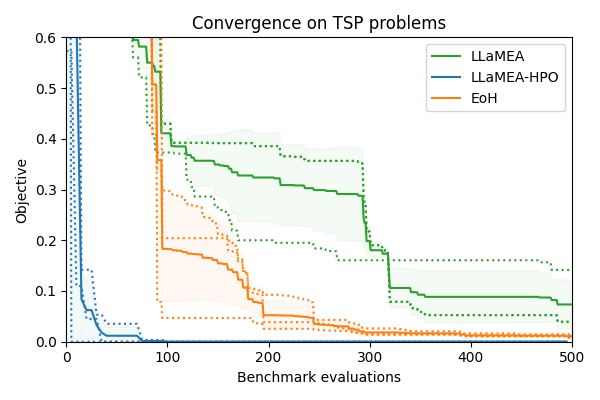}
 
 \caption{Convergence curves (lower is better) of the LLaMEA-HPO algorithm and the EoH algorithm on Traveling Sales Person Problems (TSP). Individual runs are denoted by dotted lines, the average convergence by a solid line and the standard deviation by the shaded area. The left plot shows the convergence with the number of LLM prompts used on the X-axis, and the right plot shows the convergence with the number of full benchmark evaluations used (including HPO evaluations).\label{fig:tsp}
 }
 \Description[Several lines showing the convergence of EoH and LLaMEA-HPO on TSP.]
 
\end{figure*}

In Figure \ref{fig:tsp} we can observe the convergence curves for the proposed LLaMEA-HPO against the EoH and LLaMEA baselines. In both LLM queries and complete benchmark evaluations, we can observe that the proposed approach is more efficient and finds better heuristics after only a fraction of the evaluations and prompts required by EoH or LLaMEA. Since LLaMEA uses one prompt per iteration and EoH two, we can observe a faster convergence for vanilla LLaMEA on the left plot in comparison with EoH and a slower convergence in terms of benchmark iterations in the right plot. For the TSP problem, we use a  maximum of $256$ instance evaluations for hyper-parameter-optimization, since the total training benchmark contains $64$ instances, we use $5$ evaluations of the complete benchmark in each LLM iteration.

Next we compare the performance of the final best heuristics found by the two procedures on the full test dataset of $3000$ instances. The results of this evaluation can be found in Table \ref{tab:TSP-results}. We compare in this table against the heuristics found by EoH after $100$ LLM queries and the full $2000$ LLM queries. The heuristics found by LLaMEA-HPO were found after using only $20$ LLM queries since the optimization on the training instances already converged after this number of iterations. 

\begin{table*}[ht]
\centering
\caption{Average results for algorithms on TSP20, TSP50, and TSP100 over 1000 instances per problem size. EoH-2000 uses $2000$ LLM prompts (and evaluations) \cite{fei2024eoh} and EoH-100 uses $100$ LLM prompts and evaluations. LLaMEA-HPO uses $100$ LLM prompts and $500$ evaluations.}
\begin{tabular}{l|cccccc}
\toprule
\multirow{2}{*}{Method} & \multicolumn{2}{c}{TSP20} & \multicolumn{2}{c}{TSP50} & \multicolumn{2}{c}{TSP100} \\ 
 & Gap (\%) & Time (s) & Gap (\%) & Time (s) & Gap (\%) & Time (s) \\ 
\midrule
Concorde & 0.000 & 0.010 & 0.000 & 0.051 & 0.000 & 0.224 \\ 
\midrule
AM & 0.069 & 0.038 & 0.494 & 0.124 & 2.368 & 0.356 \\
GCN & 0.035 & 0.974 & 0.884 & 3.080 & 1.880 & 6.127 \\ 
LS & 1.814 & 0.006 & 3.461 & 0.006 & 4.004 & 0.008 \\
GLS & 0.004 & 0.088 & 0.045 & 0.248 & 0.659 & 0.683 \\ 
EBGLS & 0.002 & 0.091 & 0.003 & 0.276 & 0.155 & 0.779 \\ 
KGLS & 0.000 & 1.112 & 0.000 & 3.215 & 0.035 & 7.468 \\
GNNGLS & 0.000 & 10.010 & 0.009 & 10.037 & 0.698 & 10.108 \\
NeuralGLS & 0.000 & 10.005 & 0.003 & 10.011 & 0.470 & 10.024 \\ \midrule
EoH-100-1 & 0.000 & 1.508 & 0.010 & 1.730 & 0.161 & 3.025 \\ 
EoH-100-2 & 0.000 & 1.550 & 0.000 & 2.202 & 0.031 & 4.418 \\
EoH-100-3 & 0.000 & 1.600 & 0.000 & 2.227 & 0.027 & 4.437 \\ 
\midrule
EoH-2000-1 & 0.000 & 1.854 & 0.000 & 3.925 & 0.024 & 10.085 \\ 
EoH-2000-2 & 0.000 & 1.665 & 0.000 & 2.613 & 0.027 & 5.455 \\ 
EoH-2000-3 & 0.000 & 1.590 & 0.000 & 2.244 & 0.029 & 4.549 \\ 
\midrule
\textbf{LLaMEA-HPO-1} & 0.000 & 1.709 & 0.000 & 2.678 & 0.038 & 6.271 \\ 
\textbf{LLaMEA-HPO-2} & 0.000 & 1.752 & 0.000 & 2.580 & 0.031 & 5.204 \\
\textbf{LLaMEA-HPO-3} & 0.000 & 1.964 & 0.000 & 2.744 & 0.043 & 5.253 \\ 
\bottomrule
\end{tabular}
\label{tab:TSP-results}
\end{table*}

\begin{figure*}[!htbp]
 \centering
  \includegraphics[width=\linewidth,trim=0mm 0mm 0mm 5mm,clip]{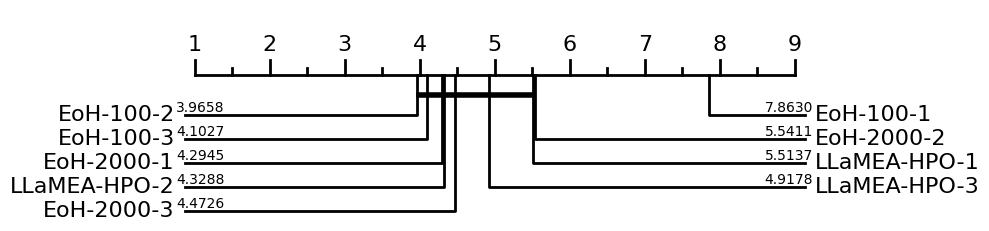}
 \caption{Critifical Difference Diagram (CDD), based on the Wilcoxon-Holm test with alpha set to $0.05$, of the different heuristics generated by EoH and LLaMEA-HPO on all Euclidean 2D TSPlib problems.\label{fig:tspcdd}
 }
 \Description[Critifical Difference Diagram of gaps on TSPlib problems for different algorithms.]
 
\end{figure*}

It is interesting to observe that only the TSP100 instances have a gap above $0.000$ for both EoH-2000 and LLaMEA-HPO discovered heuristics. Their performance is roughly on-par for these test instances. It can also be observed that most likely these instances and the training instances used in the evolutionary search procedures, are not representative and challenging enough for real world TSP problems and likely causes a form of over-fitting. We therefore also validate the generated heuristics on a large set of test instances from the commonly used TSP-lib archive \cite{reinelt1991tsplib}. The results on this wide-variety of TSP instances with sizes up to $6000$ are provided in Table \ref{tab:tsplib}. Using the Wilcoxon-holm rank for each of the generated heuristics we can calculate the critical difference diagram given in Figure \ref{fig:tspcdd}. From this diagram and from the ranks given in Table \ref{tab:tsplib}, we can conclude that most of these heuristics are not significantly different from one-another, with the exception of EoH-100-1 (the heuristic generated by EoH after $100$ evaluations on the first independent run), which is clearly worse than the others. It is interesting to note that the proposed LLaMEA-HPO algorithm only required a fraction of the LLM budget that EoH required to find similar performing heuristics.

\begin{table*}[!hbtp]
\centering
\caption{Results of EoH discovered heuristics versus LLaMEA-HPO discovered heuristics on many TSPlib problems.
}
\label{tab:tsplib}
\resizebox{0.6\textwidth}{!}{
\begin{tabular}{l|rrrrrrrrr}
\toprule

Method & \multicolumn{6}{c|}{EoH} & \multicolumn{3}{c}{LLaMEA-HPO} \\ 
\midrule
LLM prompts & \multicolumn{3}{c|}{100} & \multicolumn{3}{c|}{2000} & \multicolumn{3}{c}{20} \\ 
\midrule
a280 & 3.194 & \textbf{0.301} & 0.657 & 1.367 & 0.434 & 0.342 & 1.676 & 0.361 & 0.361 \\
bier127 & 0.396 & 0.130 & \textbf{0.010} & 0.578 & 0.130 & 0.037 & \textbf{0.010} & 0.099 & 0.048 \\
ch130 & 0.247 & \textbf{0.012} & \textbf{0.012} & \textbf{0.012} & \textbf{0.012} & 0.045 & \textbf{0.012} & 0.247 & 0.603 \\
ch150 & 1.151 & 0.239 & \textbf{0.044} & 0.370 & 0.372 & 0.227 & 0.372 & 0.372 & 0.239 \\
d1291 & 4.902 & \textbf{2.243} & 2.658 & 3.588 & 4.898 & 2.245 & 3.337 & 3.226 & 3.742 \\
d1655 & 5.043 & 3.674 & 4.519 & 3.847 & 5.043 & \textbf{3.289} & 4.050 & 4.930 & 3.498 \\
d198 & 1.707 & 0.392 & 0.458 & 0.502 & 0.419 & 0.357 & \textbf{0.296} & 0.356 & 0.305 \\
d2103 & 1.744 & 1.592 & 1.603 & 1.617 & 1.744 & 1.603 & 1.477 & \textbf{1.400} & 1.542 \\
d493 & 3.228 & \textbf{1.272} & 2.291 & 2.123 & 1.331 & 1.534 & 2.110 & 1.605 & 2.343 \\
d657 & 3.090 & 1.833 & 2.003 & 1.833 & \textbf{1.487} & 2.003 & 2.705 & 1.723 & 2.000 \\
eil101 & 3.017 & 2.131 & 2.229 & 2.322 & 1.816 & 2.138 & 1.816 & \textbf{1.782} & 2.080 \\
eil51 & 0.700 & \textbf{0.674} & \textbf{0.674} & \textbf{0.674} & \textbf{0.674} & \textbf{0.674} & \textbf{0.674} & \textbf{0.674} & \textbf{0.674} \\
eil76 & \textbf{1.184} & \textbf{1.184} & \textbf{1.184} & 1.532 & 1.641 & \textbf{1.184} & \textbf{1.184} & \textbf{1.184} & 1.477 \\
fl1400 & 11.121 & \textbf{1.541} & 4.383 & 3.458 & 7.485 & 3.344 & 10.516 & 10.611 & 8.036 \\
fl1577 & 8.019 & \textbf{4.444} & 5.582 & 6.328 & 7.227 & 5.561 & 7.227 & 6.701 & 6.591 \\
fl417 & 3.608 & \textbf{0.565} & 0.667 & 0.592 & 1.073 & 0.673 & 0.744 & 0.918 & 5.362 \\
fnl4461 & \textbf{4.319} & \textbf{4.319} & \textbf{4.319} & \textbf{4.319} & \textbf{4.319} & \textbf{4.319} & \textbf{4.319} & \textbf{4.319} & \textbf{4.319} \\
gil262 & 4.602 & 0.836 & 1.454 & \textbf{0.406} & 1.205 & 1.318 & 0.697 & 0.550 & 1.195 \\
kroA100 & \textbf{0.016} & \textbf{0.016} & \textbf{0.016} & \textbf{0.016} & \textbf{0.016} & \textbf{0.016} & \textbf{0.016} & \textbf{0.016} & \textbf{0.016} \\
kroA150 & 2.266 & 0.148 & 0.004 & \textbf{0.003} & \textbf{0.003} & 0.224 & \textbf{0.003} & 0.148 & \textbf{0.003} \\
kroA200 & 0.866 & 0.271 & 0.219 & \textbf{0.215} & 0.412 & 0.529 & 0.323 & 0.584 & 0.366 \\
kroB150 & 1.630 & 0.007 & 0.041 & \textbf{0.000} & 0.039 & 0.033 & 0.056 & 0.076 & 0.126 \\
kroB200 & 3.014 & 0.574 & 0.175 & 0.130 & 0.152 & 0.563 & 0.049 & \textbf{0.015} & 0.065 \\
kroE100 & 0.174 & 0.048 & 0.048 & 0.174 & 0.174 & \textbf{0.003} & \textbf{0.003} & \textbf{0.003} & 0.174 \\
lin318 & 3.391 & 1.359 & \textbf{0.709} & 1.109 & 1.371 & 1.390 & 2.024 & 0.890 & 1.029 \\
linhp318 & 5.101 & 3.036 & \textbf{2.375} & 2.781 & 3.048 & 3.067 & 3.712 & 2.559 & 2.700 \\
nrw1379 & 4.078 & 3.344 & \textbf{2.854} & 3.607 & 4.069 & 3.160 & 4.071 & 3.459 & 3.794 \\
p654 & 2.100 & \textbf{0.205} & 0.555 & 0.808 & 0.458 & 0.968 & 0.620 & 1.743 & 1.993 \\
pcb1173 & 5.657 & 3.345 & \textbf{3.191} & 4.356 & 5.657 & 3.303 & 4.973 & 4.418 & 4.548 \\
pcb442 & 3.831 & 1.416 & 1.453 & 1.289 & 1.316 & 1.562 & 2.227 & \textbf{0.875} & 1.802 \\
pr1002 & 3.779 & 2.792 & 2.448 & \textbf{1.884} & 3.710 & 2.508 & 3.229 & 2.551 & 2.360 \\
pr107 & 0.187 & \textbf{0.000} & \textbf{0.000} & \textbf{0.000} & \textbf{0.000} & \textbf{0.000} & \textbf{0.000} & \textbf{0.000} & \textbf{0.000} \\
pr136 & 0.734 & 0.092 & 0.112 & 0.102 & 0.092 & 0.332 & \textbf{0.009} & 0.102 & 0.014 \\
pr152 & 1.027 & \textbf{0.002} & \textbf{0.002} & 0.189 & \textbf{0.002} & \textbf{0.002} & 0.189 & \textbf{0.002} & \textbf{0.002} \\
pr226 & 0.473 & 0.078 & \textbf{0.007} & 0.170 & \textbf{0.007} & 0.088 & 0.073 & 0.073 & 0.399 \\
pr2392 & 4.550 & 4.341 & 4.238 & \textbf{3.910} & 4.238 & 4.238 & 4.262 & 4.176 & 4.253 \\
pr264 & 4.367 & 0.033 & \textbf{0.000} & \textbf{0.000} & 0.033 & \textbf{0.000} & 0.493 & 0.265 & \textbf{0.000} \\
pr299 & 4.957 & 0.603 & 0.646 & 1.579 & 0.832 & 0.662 & 1.162 & 0.529 & \textbf{0.463} \\
pr439 & 4.513 & \textbf{2.270} & 2.494 & 2.401 & 2.404 & 2.354 & 3.501 & 2.656 & 2.490 \\
pr76 & \textbf{0.000} & \textbf{0.000} & \textbf{0.000} & \textbf{0.000} & \textbf{0.000} & \textbf{0.000} & \textbf{0.000} & \textbf{0.000} & \textbf{0.000} \\
rat195 & 1.366 & 1.257 & 1.562 & 1.235 & 1.186 & 1.562 & \textbf{0.611} & 1.412 & 1.562 \\
rat575 & 3.712 & 3.482 & 2.990 & \textbf{2.387} & 3.247 & 2.691 & 3.573 & 2.770 & 3.494 \\
rat783 & 4.750 & 2.977 & 2.651 & \textbf{2.606} & 3.889 & 3.219 & 4.529 & 3.691 & 3.503 \\
rat99 & \textbf{0.681} & \textbf{0.681} & \textbf{0.681} & \textbf{0.681} & 0.732 & 0.738 & \textbf{0.681} & \textbf{0.681} & \textbf{0.681} \\
rd100 & 0.016 & \textbf{0.005} & \textbf{0.005} & \textbf{0.005} & \textbf{0.005} & \textbf{0.005} & \textbf{0.005} & \textbf{0.005} & \textbf{0.005} \\
rd400 & 2.769 & 1.035 & 0.991 & \textbf{0.930} & 1.517 & 1.102 & 2.177 & 1.413 & 0.994 \\
rl1304 & 6.496 & 2.663 & \textbf{2.489} & 5.439 & 5.675 & 3.435 & 6.349 & 4.395 & 5.766 \\
rl1323 & 4.961 & 1.997 & \textbf{1.748} & 3.286 & 4.104 & 2.108 & 4.700 & 2.342 & 3.932 \\
rl1889 & 4.104 & 3.328 & \textbf{2.926} & 3.473 & 4.075 & \textbf{2.926} & 4.104 & 3.381 & 3.828 \\
rl5915 & \textbf{3.815} & \textbf{3.815} & \textbf{3.815} & \textbf{3.815} & \textbf{3.815} & \textbf{3.815} & \textbf{3.815} & \textbf{3.815} & \textbf{3.815} \\
rl5934 & \textbf{3.670} & \textbf{3.670} & \textbf{3.670} & \textbf{3.670} & \textbf{3.670} & \textbf{3.670} & \textbf{3.670} & \textbf{3.670} & \textbf{3.670} \\
st70 & \textbf{0.313} & \textbf{0.313} & \textbf{0.313} & \textbf{0.313} & \textbf{0.313} & \textbf{0.313} & \textbf{0.313} & \textbf{0.313} & \textbf{0.313} \\
ts225 & 3.623 & \textbf{0.002} & \textbf{0.002} & \textbf{0.002} & \textbf{0.002} & \textbf{0.002} & \textbf{0.002} & \textbf{0.002} & \textbf{0.002} \\
tsp225 & \textbf{0.000} & \textbf{0.000} & \textbf{0.000} & \textbf{0.000} & 0.461 & 0.332 & \textbf{0.000} & \textbf{0.000} & \textbf{0.000} \\
u1060 & 4.562 & 2.400 & 3.139 & 3.121 & 4.086 & 2.983 & 3.997 & \textbf{2.377} & 2.702 \\
u1432 & 4.387 & 3.381 & 3.076 & 3.356 & 4.268 & \textbf{3.065} & 4.096 & 3.502 & 3.942 \\
u159 & 2.763 & \textbf{0.000} & \textbf{0.000} & \textbf{0.000} & \textbf{0.000} & \textbf{0.000} & 0.687 & \textbf{0.000} & \textbf{0.000} \\
u1817 & 5.014 & 5.006 & 5.006 & \textbf{3.772} & 5.006 & 5.006 & 4.505 & 3.853 & 4.502 \\
u2152 & 5.273 & 5.172 & 5.203 & \textbf{4.872} & 5.273 & 5.203 & 5.273 & 5.113 & 5.002 \\
u2319 & 2.639 & 2.497 & 2.324 & \textbf{2.064} & 2.568 & 2.452 & 2.639 & 2.568 & 2.603 \\
u574 & 6.241 & 2.430 & 2.579 & 1.299 & 2.367 & 2.399 & 2.108 & \textbf{1.220} & 1.708 \\
u724 & 3.897 & 2.268 & \textbf{1.615} & 1.903 & 2.889 & 2.479 & 2.597 & 1.730 & 2.337 \\
vm1084 & 3.505 & 1.581 & \textbf{1.097} & 2.224 & 2.976 & 1.176 & 3.055 & 2.000 & 2.070 \\
vm1748 & 4.293 & \textbf{2.428} & 2.706 & 3.441 & 4.078 & 2.666 & 4.110 & 2.613 & 3.350 \\
\midrule
Wilcoxon-holm rank & 7.863 & 3.966 & 4.102 & 4.295 & 5.541 & 4.473 & 5.514 & 4.329 & 4.918 \\
\bottomrule
\end{tabular}
}
\end{table*}

 
 


\section{Conclusions and Outlook}
\label{sec:conclusions}

This paper presented the hybridization of the open source LLaMEA framework \cite{vanstein2024llamealargelanguagemodel} with a specialized Hyper-Parameter Optimization (HPO) algorithm, SMAC3 \cite{smac3}, to enhance the efficiency of LLM-driven evolutionary code optimization. Our experimental results demonstrate that offloading the hyper-parameter tuning task from the LLM to an HPO procedure significantly reduces the number of LLM queries required while maintaining high solution quality. This is a crucial development in balancing computational efficiency with the computational and financial costs of querying LLMs.

One of the main findings is that LLMs are most effective when tasked with generating novel algorithmic structures and control flows rather than being utilized for simple hyper-parameter tuning. LLaMEA-HPO improves the convergence of solutions across various benchmarks, such as the Online Bin Packing, Black-Box Optimization, and the Traveling Salesperson Problem, while using significantly fewer LLM queries compared to existing approaches like EoH. By delegating hyper-parameter tuning to SMAC, our hybrid approach reduces the computational overhead associated with repeated LLM queries, without sacrificing the quality of the generated heuristics. This separation of concerns allows the LLM to focus on the creative aspects of optimization, further enhancing the diversity of solutions.

Looking ahead, future work could explore additional synergies between LLMs and more advanced HPO techniques. One promising direction is to investigate ways of dynamically adjusting the HPO budget based on the complexity of the generated algorithm, thereby further optimizing resource allocation. Additionally, expanding the framework to more diverse problem domains could further validate the generalizability of the hybrid approach. Another direction would be to investigate how the HPO-integrated procedure would affect search strategies with larger populations, such as integrating the HPO routine within EoH or within LLaMEA with larger population sizes.

In conclusion, this work underscores the importance of specializing LLM-driven optimization frameworks by offloading numerical tasks like hyper-parameter tuning to dedicated methods, ensuring the best use of computational resources while pushing the boundaries of algorithmic discovery.



\bibliographystyle{ACM-Reference-Format}
\bibliography{sample-base}


\begin{thebibliography}{32}


\ifx \showCODEN    \undefined \def \showCODEN     #1{\unskip}     \fi
\ifx \showDOI      \undefined \def \showDOI       #1{#1}\fi
\ifx \showISBNx    \undefined \def \showISBNx     #1{\unskip}     \fi
\ifx \showISBNxiii \undefined \def \showISBNxiii  #1{\unskip}     \fi
\ifx \showISSN     \undefined \def \showISSN      #1{\unskip}     \fi
\ifx \showLCCN     \undefined \def \showLCCN      #1{\unskip}     \fi
\ifx \shownote     \undefined \def \shownote      #1{#1}          \fi
\ifx \showarticletitle \undefined \def \showarticletitle #1{#1}   \fi
\ifx \showURL      \undefined \def \showURL       {\relax}        \fi
\providecommand\bibfield[2]{#2}
\providecommand\bibinfo[2]{#2}
\providecommand\natexlab[1]{#1}
\providecommand\showeprint[2][]{arXiv:#2}

\bibitem[Anonymous(2024)]%
        {anonymous_2024_13834123}
\bibfield{author}{\bibinfo{person}{Anonymous Anonymous}.} \bibinfo{year}{2024}\natexlab{}.
\newblock \bibinfo{title}{{LLaMEA-HPO: code, generated algorithms and IOH logging data.}}
\newblock
\newblock
\urldef\tempurl%
\url{https://doi.org/10.5281/zenodo.13834123}
\showDOI{\tempurl}


\bibitem[Applegate et~al\mbox{.}(2009)]%
        {applegate2009certification}
\bibfield{author}{\bibinfo{person}{David~L Applegate}, \bibinfo{person}{Robert~E Bixby}, \bibinfo{person}{Va{\v{s}}ek Chv{\'a}tal}, \bibinfo{person}{William Cook}, \bibinfo{person}{Daniel~G Espinoza}, \bibinfo{person}{Marcos Goycoolea}, {and} \bibinfo{person}{Keld Helsgaun}.} \bibinfo{year}{2009}\natexlab{}.
\newblock \showarticletitle{Certification of an optimal TSP tour through 85,900 cities}.
\newblock \bibinfo{journal}{\emph{Operations Research Letters}} \bibinfo{volume}{37}, \bibinfo{number}{1} (\bibinfo{year}{2009}), \bibinfo{pages}{11--15}.
\newblock


\bibitem[Arnold and S{\"o}rensen(2019)]%
        {arnold2019knowledge}
\bibfield{author}{\bibinfo{person}{Florian Arnold} {and} \bibinfo{person}{Kenneth S{\"o}rensen}.} \bibinfo{year}{2019}\natexlab{}.
\newblock \showarticletitle{Knowledge-guided local search for the vehicle routing problem}.
\newblock \bibinfo{journal}{\emph{Computers \& Operations Research}}  \bibinfo{volume}{105} (\bibinfo{year}{2019}), \bibinfo{pages}{32--46}.
\newblock


\bibitem[Baratchi et~al\mbox{.}(2024)]%
        {baratchi2024automated}
\bibfield{author}{\bibinfo{person}{Mitra Baratchi}, \bibinfo{person}{Can Wang}, \bibinfo{person}{Steffen Limmer}, \bibinfo{person}{Jan~N van Rijn}, \bibinfo{person}{Holger Hoos}, \bibinfo{person}{Thomas B{\"a}ck}, {and} \bibinfo{person}{Markus Olhofer}.} \bibinfo{year}{2024}\natexlab{}.
\newblock \showarticletitle{Automated machine learning: past, present and future}.
\newblock \bibinfo{journal}{\emph{Artificial Intelligence Review}} \bibinfo{volume}{57}, \bibinfo{number}{5} (\bibinfo{year}{2024}), \bibinfo{pages}{1--88}.
\newblock


\bibitem[Bischl et~al\mbox{.}(2023)]%
        {bischl2023hyperparameter}
\bibfield{author}{\bibinfo{person}{Bernd Bischl}, \bibinfo{person}{Martin Binder}, \bibinfo{person}{Michel Lang}, \bibinfo{person}{Tobias Pielok}, \bibinfo{person}{Jakob Richter}, \bibinfo{person}{Stefan Coors}, \bibinfo{person}{Janek Thomas}, \bibinfo{person}{Theresa Ullmann}, \bibinfo{person}{Marc Becker}, \bibinfo{person}{Anne-Laure Boulesteix}, {et~al\mbox{.}}} \bibinfo{year}{2023}\natexlab{}.
\newblock \showarticletitle{Hyperparameter optimization: Foundations, algorithms, best practices, and open challenges}.
\newblock \bibinfo{journal}{\emph{Wiley Interdisciplinary Reviews: Data Mining and Knowledge Discovery}} \bibinfo{volume}{13}, \bibinfo{number}{2} (\bibinfo{year}{2023}), \bibinfo{pages}{e1484}.
\newblock


\bibitem[de~Nobel et~al\mbox{.}(2024)]%
        {IOHexperimenter}
\bibfield{author}{\bibinfo{person}{Jacob de Nobel}, \bibinfo{person}{Furong Ye}, \bibinfo{person}{Diederick Vermetten}, \bibinfo{person}{Hao Wang}, \bibinfo{person}{Carola Doerr}, {and} \bibinfo{person}{Thomas B{\"{a}}ck}.} \bibinfo{year}{2024}\natexlab{}.
\newblock \showarticletitle{IOHexperimenter: Benchmarking Platform for Iterative Optimization Heuristics}.
\newblock \bibinfo{journal}{\emph{Evol. Comput.}} \bibinfo{volume}{32}, \bibinfo{number}{3} (\bibinfo{year}{2024}), \bibinfo{pages}{205--210}.
\newblock
\urldef\tempurl%
\url{https://doi.org/10.1162/EVCO\_A\_00342}
\showDOI{\tempurl}


\bibitem[Fei et~al\mbox{.}(2024)]%
        {fei2024eoh}
\bibfield{author}{\bibinfo{person}{Liu Fei}, \bibinfo{person}{Xialiang Tong}, \bibinfo{person}{Mingxuan Yuan}, \bibinfo{person}{Xi Lin}, \bibinfo{person}{Fu Luo}, \bibinfo{person}{Zhenkun Wang}, \bibinfo{person}{Zhichao Lu}, {and} \bibinfo{person}{Qingfu Zhang}.} \bibinfo{year}{2024}\natexlab{}.
\newblock \showarticletitle{Evolution of Heuristics: Towards Efficient Automatic Algorithm Design Using Large Language Model}. In \bibinfo{booktitle}{\emph{International Conference on Machine Learning (ICML)}}.
\newblock
\urldef\tempurl%
\url{https://arxiv.org/abs/2401.02051}
\showURL{%
\tempurl}


\bibitem[Feurer and Hutter(2019)]%
        {feurer2019hyperparameter}
\bibfield{author}{\bibinfo{person}{Matthias Feurer} {and} \bibinfo{person}{Frank Hutter}.} \bibinfo{year}{2019}\natexlab{}.
\newblock \showarticletitle{Hyperparameter optimization}.
\newblock \bibinfo{journal}{\emph{Automated machine learning: Methods, systems, challenges}} (\bibinfo{year}{2019}), \bibinfo{pages}{3--33}.
\newblock


\bibitem[Glickman(2012)]%
        {glickman2012example}
\bibfield{author}{\bibinfo{person}{Mark~E Glickman}.} \bibinfo{year}{2012}\natexlab{}.
\newblock \showarticletitle{Example of the Glicko-2 system}.
\newblock \bibinfo{journal}{\emph{Boston University}}  \bibinfo{volume}{28} (\bibinfo{year}{2012}).
\newblock


\bibitem[Hansen et~al\mbox{.}(2022)]%
        {hansen2022anytime}
\bibfield{author}{\bibinfo{person}{Nikolaus Hansen}, \bibinfo{person}{Anne Auger}, \bibinfo{person}{Dimo Brockhoff}, {and} \bibinfo{person}{Tea Tu{\v{s}}ar}.} \bibinfo{year}{2022}\natexlab{}.
\newblock \showarticletitle{Anytime Performance Assessment in Blackbox Optimization Benchmarking}.
\newblock \bibinfo{journal}{\emph{IEEE Transactions on Evolutionary Computation}} \bibinfo{volume}{26}, \bibinfo{number}{6} (\bibinfo{year}{2022}), \bibinfo{pages}{1293--1305}.
\newblock


\bibitem[Hansen et~al\mbox{.}(2009)]%
        {hansen2009real}
\bibfield{author}{\bibinfo{person}{Nikolaus Hansen}, \bibinfo{person}{Steffen Finck}, \bibinfo{person}{Raymond Ros}, {and} \bibinfo{person}{Anne Auger}.} \bibinfo{year}{2009}\natexlab{}.
\newblock \bibinfo{booktitle}{\emph{Real-parameter black-box optimization benchmarking 2009: Noiseless functions definitions}}.
\newblock \bibinfo{type}{{T}echnical {R}eport} RR6829. \bibinfo{institution}{INRIA}.
\newblock


\bibitem[Hoos(2012)]%
        {hoos2012automated}
\bibfield{author}{\bibinfo{person}{Holger~H Hoos}.} \bibinfo{year}{2012}\natexlab{}.
\newblock \showarticletitle{Automated algorithm configuration and parameter tuning}.
\newblock In \bibinfo{booktitle}{\emph{Autonomous search}}. \bibinfo{publisher}{Springer}, \bibinfo{pages}{37--71}.
\newblock


\bibitem[Hudson et~al\mbox{.}(2021)]%
        {hudson2021graph}
\bibfield{author}{\bibinfo{person}{Benjamin Hudson}, \bibinfo{person}{Qingbiao Li}, \bibinfo{person}{Matthew Malencia}, {and} \bibinfo{person}{Amanda Prorok}.} \bibinfo{year}{2021}\natexlab{}.
\newblock \showarticletitle{Graph neural network guided local search for the traveling salesperson problem}.
\newblock \bibinfo{journal}{\emph{arXiv preprint arXiv:2110.05291}} (\bibinfo{year}{2021}).
\newblock


\bibitem[Joshi et~al\mbox{.}(2019)]%
        {joshi2019efficient}
\bibfield{author}{\bibinfo{person}{Chaitanya~K Joshi}, \bibinfo{person}{Thomas Laurent}, {and} \bibinfo{person}{Xavier Bresson}.} \bibinfo{year}{2019}\natexlab{}.
\newblock \showarticletitle{An efficient graph convolutional network technique for the travelling salesman problem}.
\newblock \bibinfo{journal}{\emph{arXiv preprint arXiv:1906.01227}} (\bibinfo{year}{2019}).
\newblock


\bibitem[Kool et~al\mbox{.}(2018)]%
        {kool2018attention}
\bibfield{author}{\bibinfo{person}{Wouter Kool}, \bibinfo{person}{Herke Van~Hoof}, {and} \bibinfo{person}{Max Welling}.} \bibinfo{year}{2018}\natexlab{}.
\newblock \showarticletitle{Attention, learn to solve routing problems!}
\newblock \bibinfo{journal}{\emph{arXiv preprint arXiv:1803.08475}} (\bibinfo{year}{2018}).
\newblock


\bibitem[Lindauer et~al\mbox{.}(2022)]%
        {smac3}
\bibfield{author}{\bibinfo{person}{Marius Lindauer}, \bibinfo{person}{Katharina Eggensperger}, \bibinfo{person}{Matthias Feurer}, \bibinfo{person}{Andr{\'{e}} Biedenkapp}, \bibinfo{person}{Difan Deng}, \bibinfo{person}{Carolin Benjamins}, \bibinfo{person}{Tim Ruhkopf}, \bibinfo{person}{Ren{\'{e}} Sass}, {and} \bibinfo{person}{Frank Hutter}.} \bibinfo{year}{2022}\natexlab{}.
\newblock \showarticletitle{{SMAC3:} {A} Versatile Bayesian Optimization Package for Hyperparameter Optimization}.
\newblock \bibinfo{journal}{\emph{J. Mach. Learn. Res.}}  \bibinfo{volume}{23} (\bibinfo{year}{2022}), \bibinfo{pages}{54:1--54:9}.
\newblock
\urldef\tempurl%
\url{https://jmlr.org/papers/v23/21-0888.html}
\showURL{%
\tempurl}


\bibitem[Lindauer et~al\mbox{.}(2019)]%
        {configspace}
\bibfield{author}{\bibinfo{person}{M. Lindauer}, \bibinfo{person}{K. Eggensperger}, \bibinfo{person}{M. Feurer}, \bibinfo{person}{A. Biedenkapp}, \bibinfo{person}{J. Marben}, \bibinfo{person}{P. Müller}, {and} \bibinfo{person}{F. Hutter}.} \bibinfo{year}{2019}\natexlab{}.
\newblock \showarticletitle{BOAH: A Tool Suite for Multi-Fidelity Bayesian Optimization \& Analysis of Hyperparameters}.
\newblock \bibinfo{journal}{\emph{arXiv:1908.06756 {[cs.LG]}}} (\bibinfo{year}{2019}).
\newblock


\bibitem[Liu et~al\mbox{.}(2023)]%
        {liu2023algorithm}
\bibfield{author}{\bibinfo{person}{Fei Liu}, \bibinfo{person}{Xialiang Tong}, \bibinfo{person}{Mingxuan Yuan}, {and} \bibinfo{person}{Qingfu Zhang}.} \bibinfo{year}{2023}\natexlab{}.
\newblock \bibinfo{title}{Algorithm Evolution Using Large Language Model}.  (\bibinfo{year}{2023}).
\newblock
\showeprint[arxiv]{2311.15249}~[cs.NE]
\newblock
\shownote{arXiv:2311.15249}.


\bibitem[L{\'o}pez-Ib{\'a}{\~n}ez and St{\"u}tzle(2014)]%
        {lopez2014automatically}
\bibfield{author}{\bibinfo{person}{Manuel L{\'o}pez-Ib{\'a}{\~n}ez} {and} \bibinfo{person}{Thomas St{\"u}tzle}.} \bibinfo{year}{2014}\natexlab{}.
\newblock \showarticletitle{Automatically improving the anytime behaviour of optimisation algorithms}.
\newblock \bibinfo{journal}{\emph{European Journal of Operational Research}} \bibinfo{volume}{235}, \bibinfo{number}{3} (\bibinfo{year}{2014}), \bibinfo{pages}{569--582}.
\newblock


\bibitem[López-Ibáñez et~al\mbox{.}(2024)]%
        {eafecdf}
\bibfield{author}{\bibinfo{person}{Manuel López-Ibáñez}, \bibinfo{person}{Diederick Vermetten}, \bibinfo{person}{Johann Dreo}, {and} \bibinfo{person}{Carola Doerr}.} \bibinfo{year}{2024}\natexlab{}.
\newblock \bibinfo{title}{Using the Empirical Attainment Function for Analyzing Single-objective Black-box Optimization Algorithms}.  (\bibinfo{year}{2024}).
\newblock
\showeprint[arxiv]{2404.02031}~[math.OC]
\newblock
\shownote{arXiv:2404.02031}.


\bibitem[Martello and Toth(1990)]%
        {martello1990lower}
\bibfield{author}{\bibinfo{person}{Silvano Martello} {and} \bibinfo{person}{Paolo Toth}.} \bibinfo{year}{1990}\natexlab{}.
\newblock \showarticletitle{Lower bounds and reduction procedures for the bin packing problem}.
\newblock \bibinfo{journal}{\emph{Discrete applied mathematics}} \bibinfo{volume}{28}, \bibinfo{number}{1} (\bibinfo{year}{1990}), \bibinfo{pages}{59--70}.
\newblock


\bibitem[Matai et~al\mbox{.}(2010)]%
        {matai2010traveling}
\bibfield{author}{\bibinfo{person}{Rajesh Matai}, \bibinfo{person}{Surya~Prakash Singh}, {and} \bibinfo{person}{Murari~Lal Mittal}.} \bibinfo{year}{2010}\natexlab{}.
\newblock \showarticletitle{Traveling salesman problem: an overview of applications, formulations, and solution approaches}.
\newblock \bibinfo{journal}{\emph{Traveling salesman problem, theory and applications}} \bibinfo{volume}{1}, \bibinfo{number}{1} (\bibinfo{year}{2010}), \bibinfo{pages}{1--25}.
\newblock


\bibitem[OpenAI(2023a)]%
        {openai2023chatgpt4turbo}
\bibfield{author}{\bibinfo{person}{OpenAI}.} \bibinfo{year}{2023}\natexlab{a}.
\newblock \bibinfo{title}{ChatGPT-4-Turbo}.
\newblock \bibinfo{howpublished}{\url{https://platform.openai.com/docs/models/gpt-4-turbo-and-gpt-4}}.
\newblock
\newblock
\shownote{Version 2024-04-09, Accessed: 2024-05-01}.


\bibitem[OpenAI(2023b)]%
        {openai2023chatgpt4o}
\bibfield{author}{\bibinfo{person}{OpenAI}.} \bibinfo{year}{2023}\natexlab{b}.
\newblock \bibinfo{title}{ChatGPT-4o}.
\newblock \bibinfo{howpublished}{\url{https://platform.openai.com/docs/models/gpt-4o}}.
\newblock
\newblock
\shownote{Version: 2024-05-13, Accessed: 2024-05-14}.


\bibitem[Reinelt(1991)]%
        {reinelt1991tsplib}
\bibfield{author}{\bibinfo{person}{Gerhard Reinelt}.} \bibinfo{year}{1991}\natexlab{}.
\newblock \showarticletitle{TSPLIB—A traveling salesman problem library}.
\newblock \bibinfo{journal}{\emph{ORSA journal on computing}} \bibinfo{volume}{3}, \bibinfo{number}{4} (\bibinfo{year}{1991}), \bibinfo{pages}{376--384}.
\newblock


\bibitem[Romera-Paredes et~al\mbox{.}(2024)]%
        {FunSearch2024}
\bibfield{author}{\bibinfo{person}{Bernardino Romera-Paredes}, \bibinfo{person}{Mohammadamin Barekatain}, \bibinfo{person}{Alexander Novikov}, \bibinfo{person}{Matej Balog}, \bibinfo{person}{M~Pawan Kumar}, \bibinfo{person}{Emilien Dupont}, \bibinfo{person}{Francisco~JR Ruiz}, \bibinfo{person}{Jordan~S Ellenberg}, \bibinfo{person}{Pengming Wang}, \bibinfo{person}{Omar Fawzi}, \bibinfo{person}{Pushmeet Kohli}, {and} \bibinfo{person}{Alhussein Fawzi}.} \bibinfo{year}{2024}\natexlab{}.
\newblock \showarticletitle{Mathematical discoveries from program search with large language models}.
\newblock \bibinfo{journal}{\emph{Nature}}  \bibinfo{volume}{625} (\bibinfo{date}{01} \bibinfo{year}{2024}), \bibinfo{pages}{468--475}.
\newblock
Issue 7995.


\bibitem[Seiden(2002)]%
        {seiden2002online}
\bibfield{author}{\bibinfo{person}{Steven~S Seiden}.} \bibinfo{year}{2002}\natexlab{}.
\newblock \showarticletitle{On the online bin packing problem}.
\newblock \bibinfo{journal}{\emph{Journal of the ACM (JACM)}} \bibinfo{volume}{49}, \bibinfo{number}{5} (\bibinfo{year}{2002}), \bibinfo{pages}{640--671}.
\newblock


\bibitem[Shi et~al\mbox{.}(2018)]%
        {shi2018eb}
\bibfield{author}{\bibinfo{person}{Jialong Shi}, \bibinfo{person}{Qingfu Zhang}, {and} \bibinfo{person}{Edward Tsang}.} \bibinfo{year}{2018}\natexlab{}.
\newblock \showarticletitle{EB-GLS: an improved guided local search based on the big valley structure}.
\newblock \bibinfo{journal}{\emph{Memetic computing}}  \bibinfo{volume}{10} (\bibinfo{year}{2018}), \bibinfo{pages}{333--350}.
\newblock


\bibitem[Sui et~al\mbox{.}(2024)]%
        {sui2024neuralgls}
\bibfield{author}{\bibinfo{person}{Jingyan Sui}, \bibinfo{person}{Shizhe Ding}, \bibinfo{person}{Boyang Xia}, \bibinfo{person}{Ruizhi Liu}, {and} \bibinfo{person}{Dongbo Bu}.} \bibinfo{year}{2024}\natexlab{}.
\newblock \showarticletitle{NeuralGLS: learning to guide local search with graph convolutional network for the traveling salesman problem}.
\newblock \bibinfo{journal}{\emph{Neural Computing and Applications}} \bibinfo{volume}{36}, \bibinfo{number}{17} (\bibinfo{year}{2024}), \bibinfo{pages}{9687--9706}.
\newblock


\bibitem[van Stein and Bäck(2024)]%
        {vanstein2024llamealargelanguagemodel}
\bibfield{author}{\bibinfo{person}{Niki van Stein} {and} \bibinfo{person}{Thomas Bäck}.} \bibinfo{year}{2024}\natexlab{}.
\newblock \bibinfo{title}{LLaMEA: A Large Language Model Evolutionary Algorithm for Automatically Generating Metaheuristics}.
\newblock
\newblock
\showeprint[arxiv]{2405.20132}~[cs.NE]
\urldef\tempurl%
\url{https://arxiv.org/abs/2405.20132}
\showURL{%
\tempurl}


\bibitem[Voudouris and Tsang(1999)]%
        {voudouris1999guided}
\bibfield{author}{\bibinfo{person}{Christos Voudouris} {and} \bibinfo{person}{Edward Tsang}.} \bibinfo{year}{1999}\natexlab{}.
\newblock \showarticletitle{Guided local search and its application to the traveling salesman problem}.
\newblock \bibinfo{journal}{\emph{European journal of operational research}} \bibinfo{volume}{113}, \bibinfo{number}{2} (\bibinfo{year}{1999}), \bibinfo{pages}{469--499}.
\newblock


\bibitem[Zhang et~al\mbox{.}(2024)]%
        {zhang2024understanding}
\bibfield{author}{\bibinfo{person}{Rui Zhang}, \bibinfo{person}{Fei Liu}, \bibinfo{person}{Xi Lin}, \bibinfo{person}{Zhenkun Wang}, \bibinfo{person}{Zhichao Lu}, {and} \bibinfo{person}{Qingfu Zhang}.} \bibinfo{year}{2024}\natexlab{}.
\newblock \showarticletitle{Understanding the Importance of Evolutionary Search in Automated Heuristic Design with Large Language Models}. In \bibinfo{booktitle}{\emph{International Conference on Parallel Problem Solving from Nature}}. Springer, \bibinfo{pages}{185--202}.
\newblock


\end{thebibliography}

\FloatBarrier
\appendix

\onecolumn

\newpage

\section{Benchmark Dependent Prompts}

Below are the specific task dependent prompts used by LLaMEA-HPO. These prompts are as close as possible to the prompts used by EoH and LLaMEA, the only difference is the format specification of the output which requires different instructions between the frameworks, and the fact that EoH and vanilla LLaMEA cannot handle the configuration space generation required for performing hyper-parameter optimization. All prompts used and all source code are available at our Zenodo repository for further details \cite{anonymous_2024_13834123}.

\medskip

\begin{tiny}
\begin{Verbatim}[breaklines=true,frame=single, label=Detailed Task Prompt for Online Bin Packing]
I need help designing a novel score function that scoring a set of bins to assign an item.
In each step, the item will be assigned to the bin with the maximum score. If the rest capacity of a bin equals the maximum capacity, it will not be used. The final goal is to minimize the number of used bins.

The heuristic algorithm class should contain two functions an "__init__()" function containing any hyper-parameters that can be optimmized, and a "score(self, item, bins)" function, which gives back the 'scores'.
'item' and 'bins' are the size of the current item and the rest capacities of feasible bins, which are larger than the item size.
The output named 'scores' is the scores for the bins for assignment.
Note that 'item' is of type int, while 'bins' and 'scores' are both Numpy arrays. The novel function should be sufficiently complex in order to achieve better performance. It is important to ensure self-consistency.

An example baseline heuristic that we should improve and to show the structure is as follows:
```python
import numpy as np

class Sample:
    def __init__(self, s1=1.0, s2=100):
        self.s1 = s1
        self.s2 = s2

    def score(self, item, bins):
        scores = items - bins
    return scores
```

In addition, any hyper-parameters the algorithm uses will be optimized by SMAC, for this, provide a Configuration space as Python dictionary (without the item and bins parameters) and include all hyper-parameters to be optimized in the __init__ function header.
An example configuration space is as follows:

```python
{
    "float_parameter": (0.1, 1.5),
    "int_parameter": (2, 10), 
    "categoral_parameter": ["mouse", "cat", "dog"]
}
```

Give an excellent and novel heuristic including its configuration space to solve this task and also give it a name. Give the response in the format:
# Name: <name>
# Code: <code>
# Space: <configuration_space>
\end{Verbatim}
\end{tiny}

\medskip

\begin{tiny}
\begin{Verbatim}[breaklines=true,frame=single, label=Detailed Task Prompt for Online Bin Packing] 
Task: Given an edge distance matrix and a local optimal route, please help me design a strategy to update the distance matrix to avoid being trapped in the local optimum with the final goal of finding a tour with minimized distance (TSP problem).
You should create an algorithm for me to update the edge distance matrix.
Provide the Python code for the new strategy. The code is a Python class that should contain two functions an "__init__()" function containing any hyper-parameters that can be optimmized, and a 
function called 'update_edge_distance(self, edge_distance, local_opt_tour, edge_n_used)' that takes three inputs, and outputs the 'updated_edge_distance', 
where 'local_opt_tour' includes the local optimal tour of IDs, 'edge_distance' and 'edge_n_used' are matrixes, 'edge_n_used' includes the number of each edge used during permutation. 
All are Numpy arrays. 
The novel function should be sufficiently complex in order to achieve better performance. It is important to ensure self-consistency.

An example heuristic to show the structure is as follows.
```python
import numpy as np

class Sample:
    def __init__(self, param1, param2):
        self.param1 = param1
        self.param2 = param2

    def update_edge_distance(self, edge_distance, local_opt_tour, edge_n_used):
        # code here
        return updated_edge_distance
```

In addition, any hyper-parameters the algorithm used will be optimized by SMAC, for this, provide a Configuration space as Python dictionary (without the edge_distance, local_opt_tour, edge_n_used parameters) and include all hyper-parameters to be optimized in the __init__ function header.
An example configuration space is as follows:

```python
{
    "float_parameter": (0.1, 1.5),
    "int_parameter": (2, 10), 
    "categoral_parameter": ["mouse", "cat", "dog"]
}
```

Give an excellent and novel heuristic including its configuration space to solve this task and also give it a name. Give the response in the format:
# Name: <name>
# Code: <code>
# Space: <configuration_space>
\end{Verbatim}
\end{tiny}

\medskip

\begin{tiny}
\begin{Verbatim}[breaklines=true,frame=single, label=Detailed Task Prompt for Online Bin Packing] 
Your task is to design novel metaheuristic algorithms to solve black box optimization problems.
The optimization algorithm should handle a wide range of tasks, which is evaluated on a test suite of noiseless functions. Your task is to write the optimization algorithm in Python code. The code should contain an `__init__(self, budget, dim)` function with optional additional arguments and the function `def __call__(self, func)`, which should optimize the black box function `func` using `self.budget` function evaluations.
The func() can only be called as many times as the budget allows, not more. Each of the optimization functions has a search space between -5.0 (lower bound) and 5.0 (upper bound). The dimensionality can be varied.
An example of such code (a simple random search), is as follows:
```python
import numpy as np

class RandomSearch:
    def __init__(self, budget=10000, dim=10):
        self.budget = budget
        self.dim = dim

    def __call__(self, func):
        self.f_opt = np.Inf
        self.x_opt = None
        for i in range(self.budget):
            x = np.random.uniform(func.bounds.lb, func.bounds.ub)
            
            f = func(x)
            if f < self.f_opt:
                self.f_opt = f
                self.x_opt = x
            
        return self.f_opt, self.x_opt
```

In addition, any hyper-parameters the algorithm uses will be optimized by SMAC, for this, provide a Configuration space as Python dictionary (without the dim and budget parameters) and include all hyper-parameters in the __init__ function header.
An example configuration space is as follows:

```python
{
    "float_parameter": (0.1, 1.5),
    "int_parameter": (2, 10), 
    "categoral_parameter": ["mouse", "cat", "dog"]
}
```

Give an excellent and novel heuristic algorithm including its configuration space to solve this task and also give it a name. Give the response in the format:
# Name: <name>
# Code: <code>
# Space: <configuration_space>
\end{Verbatim}
\end{tiny}
\newpage

\section{BBOB Detailed Convergence Curves for $5d$}

\begin{figure}[H]
 \centering
  \includegraphics[width=\linewidth,trim=0mm 0mm 0mm 0mm,clip]{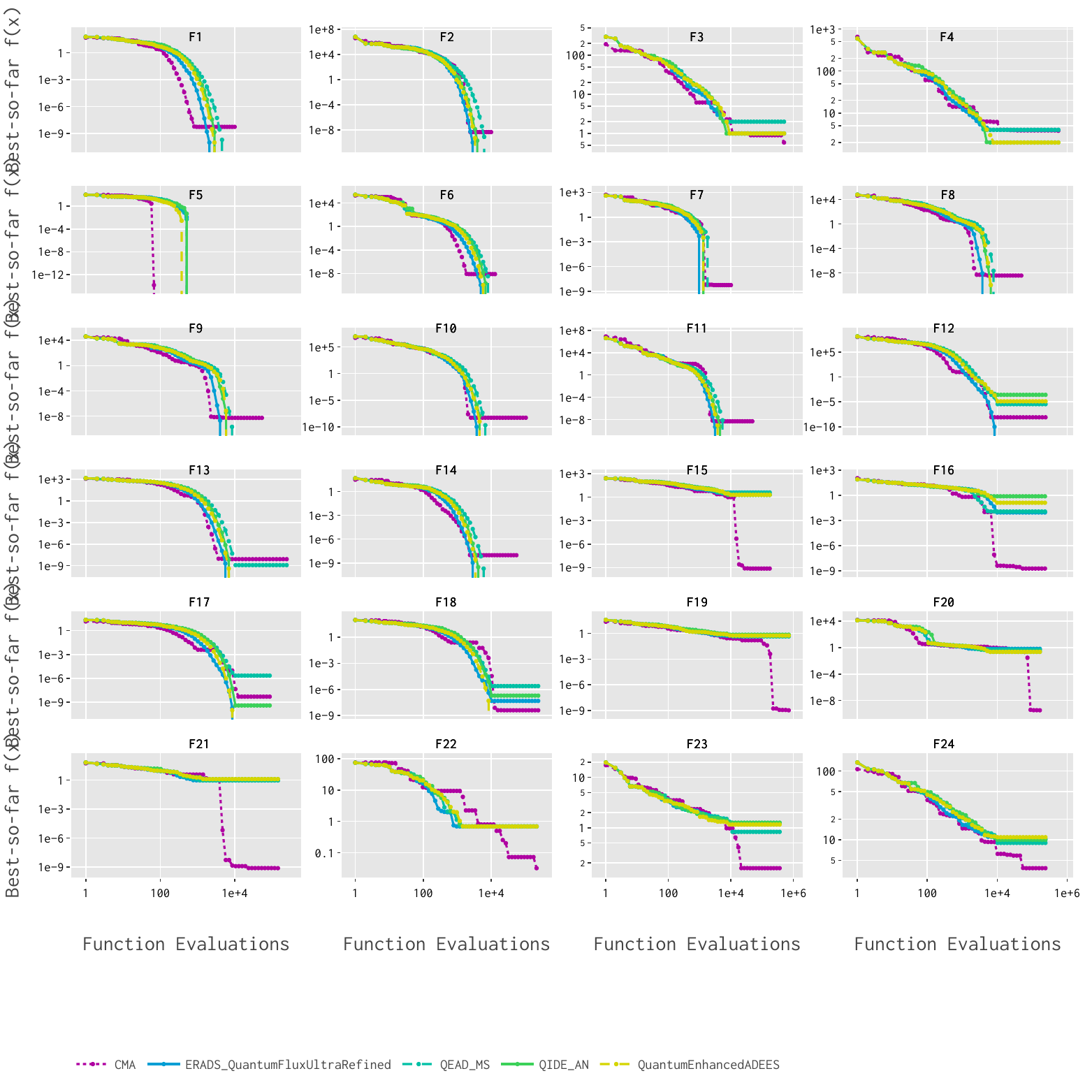}
 \caption{Convergence curves in 5d for each BBOB benchmark function for the original LLaMEA's generated best black-box optimization algorithm (ERADS) versus optimization algorithms generated by LLaMEA-HPO.\label{fig:eradsFCE}
 }
 \Description[Several lines showing the convergence of different algorithms on BBOB in 5d]
 
\end{figure}

Detailed convergence curves on each BBOB function are given in Figure \ref{fig:eradsFCE} for the LLaMEA-HPO generated algorithms versus the original ``ERADS\_QuantumFluxUltraEnhanced'' algorithm from \cite{vanstein2024llamealargelanguagemodel}.

\section{TSP Algorithms}

In the following section we provide the source codes of the TSP heuristics compared in Table \ref{tab:tsplib} of LLaMEA-HPO and EoH with the full $2000$ LLM evaluations budget. It is interesting to observe the different complexity of codes and the fact that LLaMEA-HPO can produce more complex classes with internal memory mechanisms, which EoH is unable to produce due to their limitation of generating single functions.

\begin{listing}[H]
\tiny
\begin{minted}{python}
import numpy as np
# best config = {'alpha': 0.4819640658796, 'beta': 0.1594372905791, 'delta': 0.5715060189366, 
# 'epsilon': 1.6157230922952, 'eta': 0.8430815145373, 'gamma': 0.8607733716257, 
# 'zeta': 0.0325028870814}

class HeuristicAugmentedPenaltiesTSP:
    def __init__(self, alpha, beta, gamma, delta, epsilon, zeta, eta):
        """
        :param alpha: A parameter controlling the penalty strength based on the usage frequency of edges.
        :param beta: A decay parameter that reduces the penalty strength over iterations.
        :param gamma: A parameter for increasing exploration by adding penalties to frequently used edges in the local optimal tour.
        :param delta: A parameter for dynamically adjusting penalties based on the variance of edge usage.
        :param epsilon: A parameter for boosting the penalty for the most frequently used edges.
        :param zeta: A parameter for introducing randomness to escape local optima.
        :param eta: A parameter for selective penalization of the critical edges forming cycles.
        """
        self.alpha = alpha
        self.beta = beta
        self.gamma = gamma
        self.delta = delta
        self.epsilon = epsilon
        self.zeta = zeta
        self.eta = eta

    def update_edge_distance(self, edge_distance, local_opt_tour, edge_n_used):
        """
        Updates the edge distances to avoid being trapped in the local optimum.
        
        :param edge_distance: 2D numpy array representing the edge distances.
        :param local_opt_tour: 1D numpy array representing the current local optimal tour.
        :param edge_n_used: 2D numpy array representing the number of times each edge has been used.
        :return: 2D numpy array representing the updated edge distances.
        """
        n = edge_distance.shape[0]
        updated_edge_distance = np.copy(edge_distance)

        # Compute the variance of edge usage for dynamic adjustment
        edge_usage_variance = np.var(edge_n_used)

        # Determine the most frequently used edges
        max_usage = np.max(edge_n_used)

        # Apply penalties to edges based on their usage frequency and the local optimal tour
        for i in range(len(local_opt_tour) - 1):
            u, v = local_opt_tour[i], local_opt_tour[i + 1]
            usage_penalty = self.alpha * (edge_n_used[u, v] + self.gamma) + self.delta * edge_usage_variance

            # Boost penalty for the most frequently used edges
            if edge_n_used[u, v] == max_usage:
                usage_penalty *= self.epsilon

            updated_edge_distance[u, v] += usage_penalty
            updated_edge_distance[v, u] += usage_penalty  # Since the matrix is symmetric

        # Detect and penalize critical edges forming cycles
        for i in range(len(local_opt_tour) - 1):
            for j in range(i + 2, len(local_opt_tour) - 1):
                if local_opt_tour[i] == local_opt_tour[j + 1] and local_opt_tour[i + 1] == local_opt_tour[j]:
                    critical_edge_penalty = self.eta * edge_n_used[local_opt_tour[i], local_opt_tour[i + 1]]
                    updated_edge_distance[local_opt_tour[i], local_opt_tour[i + 1]] += critical_edge_penalty
                    updated_edge_distance[local_opt_tour[i + 1], local_opt_tour[i]] += critical_edge_penalty

        # Apply a decay to the previous penalties to allow exploration of other paths
        decayed_penalty = np.exp(-self.beta * edge_n_used)
        updated_edge_distance = updated_edge_distance * decayed_penalty

        # Introduce randomness to help escape local optima
        random_penalty = np.random.rand(n, n) * self.zeta
        updated_edge_distance += random_penalty

        return updated_edge_distance
\end{minted}
\caption{TSP heuristic generated by LLaMEA-HPO-1}
\label{listing:1}
\end{listing}

\begin{listing}[H]
\tiny
\begin{minted}{python}
import numpy as np
# best config = {'alpha': 1.1058783142151, 'beta': 0.1964654028153, 'decay_rate': 0.2105952695588,
# 'edge_reward': 0.9816288003968, 'exploration_weight': 0.083229970681,  'feedback_weight': 0.7705326445771, 
# 'gamma': 0.0305881625952, 'max_penalty': 1.0608059932778, 'memory_decay': 0.5605192650336, 
# 'memory_weight': 0.6059271080345, 'min_penalty': 0.8272987030459, 'penalty_scaling': 0.0898495276382}
class RefinedDynamicEdgeMemoryPenaltyTSP:
    def __init__(self, alpha=1.0, beta=0.5, gamma=0.1, max_penalty=10.0, min_penalty=0.1, decay_rate=0.1, feedback_weight=0.5, 
        penalty_scaling=0.01, memory_weight=0.9, exploration_weight=0.05, memory_decay=0.95, edge_reward=1.0):
        # skipped init code due to space limits
        self.memory = None

    def initialize_memory(self, n):
        self.memory = np.zeros((n, n))

    def update_edge_distance(self, edge_distance, local_opt_tour, edge_n_used):
        """
        Updates the edge distances to avoid being trapped in the local optimum.
        
        :param edge_distance: 2D numpy array representing the edge distances.
        :param local_opt_tour: 1D numpy array representing the current local optimal tour.
        :param edge_n_used: 2D numpy array representing the number of times each edge has been used.
        :return: 2D numpy array representing the updated edge distances.
        """
        n = edge_distance.shape[0]
        if self.memory is None:
            self.initialize_memory(n)
        
        updated_edge_distance = np.copy(edge_distance)
        penalty_matrix = np.zeros_like(edge_distance)
        # Apply penalties to edges based on their usage frequency and the local optimal tour
        for i in range(len(local_opt_tour) - 1):
            u, v = local_opt_tour[i], local_opt_tour[i + 1]
            usage_penalty = self.alpha * np.log(1 + edge_n_used[u, v])  # Logarithmic penalty to smoothen the impact
            # Cap the penalty to avoid excessive increase
            usage_penalty = min(max(usage_penalty, self.min_penalty), self.max_penalty)
            penalty_matrix[u, v] += usage_penalty
            penalty_matrix[v, u] += usage_penalty  # Since the matrix is symmetric
        # Apply a non-linear decay to the previous penalties
        penalty_matrix *= np.exp(-self.beta * np.sqrt(edge_n_used))
        # Dynamic exploration factor to further incentivize exploration
        exploration_factor = self.gamma * (np.max(edge_n_used) - edge_n_used)
        penalty_matrix += self.exploration_weight * exploration_factor
        # Normalize the updated edge distances to maintain consistency
        min_distance = np.min(penalty_matrix[np.nonzero(penalty_matrix)])
        max_distance = np.max(penalty_matrix)
        normalized_penalty_matrix = (penalty_matrix - min_distance) / (max_distance - min_distance + 1e-6)
        updated_edge_distance += normalized_penalty_matrix
        # Apply overall decay to penalties to avoid runaway increases over long iterations
        updated_edge_distance *= (1 - self.decay_rate)
        # Feedback mechanism based on the overall quality of the tour
        tour_length = sum(edge_distance[local_opt_tour[i], local_opt_tour[i + 1]] for i in range(len(local_opt_tour) - 1))
        feedback_penalty = self.feedback_weight * (tour_length - np.mean(edge_distance))
        if feedback_penalty > 0:
            penalty_matrix += feedback_penalty
        # Reward mechanism to reinforce good edges
        reward_matrix = np.zeros_like(edge_distance)
        for i in range(len(local_opt_tour) - 1):
            u, v = local_opt_tour[i], local_opt_tour[i + 1]
            reward_matrix[u, v] += self.edge_reward
            reward_matrix[v, u] += self.edge_reward  # Since the matrix is symmetric
        penalty_matrix -= reward_matrix
        # Update memory with decay
        if self.memory is not None:
            self.memory = self.memory * self.memory_decay + penalty_matrix

        # Apply historical memory to the updated edge distances
        if self.memory is not None:
            updated_edge_distance += self.penalty_scaling * self.memory

        return updated_edge_distance
\end{minted}
\caption{TSP heuristic generated by LLaMEA-HPO-2}
\label{listing:2}
\end{listing}

\begin{listing}[H]
\tiny
\begin{minted}{python}
import numpy as np
# best config = {'alpha': 2.8911497606854, 'beta': 0.6327692118248, 'decay_rate': 0.7696552437427, 'elite_frac': 0.0439335341365, 
#'exploration_weight': 0.0252923145741, 'feedback_weight': 0.0766586671562, 'gamma': 2.4691103820661, 'max_penalty': 8.1535746868849, 
#'memory_weight': 0.0513291250079, 'min_penalty': 0.0532730015106, 'penalty_scaling': 4.3408972140962}

class RefinedDynamicEdgePenaltyPlusTSP:
    def __init__(self, alpha, beta, gamma, max_penalty, min_penalty, decay_rate, elite_frac, feedback_weight, 
        penalty_scaling, memory_weight, exploration_weight):
        # skipped init code
        self.memory = None

    def initialize_memory(self, n):
        self.memory = np.zeros((n, n))

    def update_edge_distance(self, edge_distance, local_opt_tour, edge_n_used):
        """
        Updates the edge distances to avoid being trapped in the local optimum.
        
        :param edge_distance: 2D numpy array representing the edge distances.
        :param local_opt_tour: 1D numpy array representing the current local optimal tour.
        :param edge_n_used: 2D numpy array representing the number of times each edge has been used.
        :return: 2D numpy array representing the updated edge distances.
        """
        n = edge_distance.shape[0]
        if self.memory is None:
            self.initialize_memory(n)
        
        updated_edge_distance = np.copy(edge_distance)
        penalty_matrix = np.zeros_like(edge_distance)

        # Apply penalties to edges based on their usage frequency and the local optimal tour
        for i in range(len(local_opt_tour) - 1):
            u, v = local_opt_tour[i], local_opt_tour[i + 1]
            usage_penalty = self.alpha * np.log(1 + edge_n_used[u, v])  # Logarithmic penalty to smoothen the impact

            # Cap the penalty to avoid excessive increase
            usage_penalty = min(max(usage_penalty, self.min_penalty), self.max_penalty)

            penalty_matrix[u, v] += usage_penalty
            penalty_matrix[v, u] += usage_penalty  # Since the matrix is symmetric

        # Apply a non-linear decay to the previous penalties
        penalty_matrix *= np.exp(-self.beta * np.sqrt(edge_n_used))

        # Dynamic exploration factor to further incentivize exploration
        exploration_factor = self.gamma * (np.max(edge_n_used) - edge_n_used)
        penalty_matrix += self.exploration_weight * exploration_factor

        # Normalize the updated edge distances to maintain consistency
        min_distance = np.min(penalty_matrix[np.nonzero(penalty_matrix)])
        max_distance = np.max(penalty_matrix)
        normalized_penalty_matrix = (penalty_matrix - min_distance) / (max_distance - min_distance + 1e-6)

        updated_edge_distance += normalized_penalty_matrix

        # Apply overall decay to penalties to avoid runaway increases over long iterations
        updated_edge_distance *= (1 - self.decay_rate)

        # Preserve a fraction of edges as elite to avoid excessive penalties
        flat_distances = updated_edge_distance.flatten()
        elite_threshold = np.percentile(flat_distances, self.elite_frac * 100)
        elite_mask = updated_edge_distance < elite_threshold
        updated_edge_distance[elite_mask] = edge_distance[elite_mask]

        # Feedback mechanism based on the overall quality of the tour
        tour_length = sum(edge_distance[local_opt_tour[i], local_opt_tour[i + 1]] for i in range(len(local_opt_tour) - 1))
        feedback_penalty = self.feedback_weight * (tour_length - np.mean(edge_distance))
        if feedback_penalty > 0:
            penalty_matrix += feedback_penalty

        # Update memory with decay
        if self.memory is not None:
            self.memory = self.memory_weight * self.memory + penalty_matrix

        # Apply historical memory to the updated edge distances
        if self.memory is not None:
            updated_edge_distance += self.penalty_scaling * self.memory

        return updated_edge_distance
\end{minted}
\caption{TSP heuristic generated by LLaMEA-HPO-3}
\label{listing:3}
\end{listing}

\begin{listing}[H]
\tiny
\begin{minted}{python}
import numpy as np

def update_edge_distance(edge_distance, local_opt_tour, edge_n_used):
    updated_edge_distance = np.copy(edge_distance)
    
    edge_count = np.zeros_like(edge_distance)
    for i in range(len(local_opt_tour) - 1):
        start = local_opt_tour[i]
        end = local_opt_tour[i + 1]
        edge_count[start][end] += 1
        edge_count[end][start] += 1
    
    edge_n_used_max = np.max(edge_n_used)
    decay_factor = 0.1
    mean_distance = np.mean(edge_distance)
    for i in range(edge_distance.shape[0]):
        for j in range(edge_distance.shape[1]):
            if edge_count[i][j] > 0:
                noise_factor = (np.random.uniform(0.7, 1.3) / edge_count[i][j]) 
                    + (edge_distance[i][j] / mean_distance) - (0.3 / edge_n_used_max) * edge_n_used[i][j]
                updated_edge_distance[i][j] += noise_factor * (1 + edge_count[i][j]) 
                    - decay_factor * updated_edge_distance[i][j]

    return updated_edge_distance
\end{minted}
\caption{TSP heuristic generated by EoH (EoH-2000-1)}
\label{listing:EoH1}
\end{listing}

\begin{listing}[H]
\tiny
\begin{minted}{python}
import numpy as np

def update_edge_distance(edge_distance, local_opt_tour, edge_n_used):
    updated_edge_distance = edge_distance.copy()
    for i in range(len(local_opt_tour)-1):
        edge = (local_opt_tour[i], local_opt_tour[i+1])
        edge_n_used_normalized = edge_n_used[edge] / np.max(edge_n_used)
        edge_distance_increase = 5.0 + 0.3 / np.power(edge_n_used_normalized + 5, 0.7)  
        # Different parameter setting, increase factor of 0.3 with power normalization
        updated_edge_distance[edge] += edge_distance_increase
        updated_edge_distance[edge[::-1]] += edge_distance_increase  # Update the reverse edge as well
    return updated_edge_distance
\end{minted}
\caption{TSP heuristic generated by EoH (EoH-2000-2)}
\label{listing:EoH2}
\end{listing}

\begin{listing}[H]
\tiny
\begin{minted}{python}
import numpy as np
def update_edge_distance(edge_distance, local_opt_tour, edge_n_used):
    updated_edge_distance = np.copy(edge_distance)
    max_n_used = np.max(edge_n_used)
    penalty_factor = 0.6 * (max_n_used - edge_n_used)**1.2
    
    for i in range(len(local_opt_tour)-1):
        edge_i = local_opt_tour[i]
        edge_j = local_opt_tour[i+1]
        
        updated_edge_distance[edge_i][edge_j] += penalty_factor[edge_i][edge_j]
        updated_edge_distance[edge_j][edge_i] = updated_edge_distance[edge_i][edge_j] 
        # symmetrical matrix
        
    return updated_edge_distance
\end{minted}
\caption{TSP heuristic generated by EoH (EoH-2000-3)}
\label{listing:EoH3}
\end{listing}

\end{document}